\documentclass[preprint]{elsarticle}

\makeatletter
\def\ps@pprintTitle{%
  \let\@oddfoot\@empty
  \let\@evenfoot\@empty
}
\makeatother

\usepackage{setspace}

\usepackage[a4paper, left=2.5cm, right=2.5cm, top=3cm, bottom=3cm]{geometry}

\usepackage{float} 

\usepackage{amssymb}
\usepackage{amsmath}
\usepackage{orcidlink}
\usepackage[utf8]{inputenc}
\DeclareUnicodeCharacter{2194}{\leftrightarrow}

\usepackage{pdflscape}
\usepackage{booktabs}  
\usepackage{pifont}    
\usepackage{amsmath,amssymb,amsfonts}
\usepackage{algorithmic}
\usepackage{graphicx}   
\usepackage{etoolbox}   

\usepackage{textcomp}
\def\BibTeX{{\rm B\kern-.05em{\sc i\kern-.025em b}\kern-.08em
    T\kern-.1667em\lower.7ex\hbox{E}\kern-.125emX}}

\usepackage{multicol}
\usepackage{float}
\usepackage{soul}
\usepackage{algorithmic}
\usepackage{array}
\usepackage{url}

\usepackage{makecell}
\usepackage{graphicx}
\usepackage{booktabs}
\usepackage{multirow}
\usepackage{hyperref}

\usepackage{listings}
\usepackage{xcolor}
\usepackage{tikz}
\usetikzlibrary{positioning, shapes.geometric, calc}

\definecolor{codegreen}{rgb}{0,0.6,0}
\definecolor{codegray}{rgb}{0.97,0.97,0.97}
\definecolor{codepurple}{rgb}{0.58,0,0.82}
\definecolor{backcolour}{rgb}{0.95,0.95,0.92}

\lstdefinestyle{mystyle}{
	backgroundcolor=\color{codegray},   
	commentstyle=\color{codegreen},
	keywordstyle=\color{black},
	numberstyle=\tiny\color{codegray},
	stringstyle=\color{black},
	basicstyle=\ttfamily,
	breakatwhitespace=false,         
	breaklines=true,                 
	captionpos=b,                    
	keepspaces=true,                 
	numbersep=5pt,                  
	showspaces=false,                
	showstringspaces=false,
	showtabs=false,                  
	tabsize=2
}

\newcolumntype{M}[1]{>{\centering\arraybackslash}m{#1}}

\begin{document}

\begin{frontmatter}



\title{Causal Transfer in Medical Image Analysis}



\author[Exeter]{Mohammed M. Abdelsamea\corref{cor1}}
\author[Exeter]{Daniel Tweneboah Anyimadu}
\author[Alex]{Tasneem Selim}
\author[Exeter]{Saif Alzubi}
\author[Exeter]{Lei Zhang}
\author[Exeter,UCL]{Ahmed Karam Eldaly}
\author[Exeter]{Xujiong Ye}

\cortext[cor1]{Corresponding author: Mohammed M. Abdelsamea, email: m.abdelsamea@exeter.ac.uk}

\affiliation[Exeter]{organization={Department of Computer Science, University of Exeter},
            city={Exeter},
            country={United Kingdom}}

\affiliation[Alex]{organization={Department of Mathematics and Computer Science, Faculty of Science, Alexandria University},
            city={Alexandria},
            country={Egypt}}  

\affiliation[UCL]{organization={Hawkes Institute, Department of Computer Science, University College London},
            city={London},
            country={United Kingdom}}

\begin{abstract}
Medical imaging models frequently fail when deployed across hospitals, scanners, populations, or imaging protocols due to domain shift, limiting their clinical reliability. While transfer learning and domain adaptation address such shifts statistically, they often rely on spurious correlations that break under changing conditions. On the other hand, causal inference provides a principled way to identify invariant mechanisms that remain stable across environments. This survey introduces and systematises Causal Transfer Learning \emph{(CTL)} for medical image analysis, framing domain shift through a causal lens and explicitly linking it to four categories of distribution shift: covariate, label, conditional, and concept shift. This perspective clarifies when and why standard transfer learning methods succeed or fail across clinical environments. This paradigm integrates causal reasoning with cross-domain representation learning to enable robust and generalisable clinical AI. We frame domain shift as a causal problem and analyse how structural causal models, invariant risk minimisation, causal discovery, and counterfactual reasoning can be embedded within transfer learning pipelines. We survey studies spanning classification, segmentation, reconstruction, enhancement, anomaly detection, and multimodal imaging, and organise them by task, shift type, and underlying causal assumptions. This results in a unified taxonomy that connects causal frameworks with transfer mechanisms and explicitly relates each methodological class to the type of distribution shift it is designed to handle. We further summarise datasets, benchmarks, and empirical gains, highlighting when and why causal transfer outperforms correlation-based domain adaptation. Finally, we discuss how CTL supports fairness, robustness, and trustworthy deployment in multi-institutional and federated settings, and outline open challenges and research directions for clinically reliable medical imaging AI.
\end{abstract}




\end{frontmatter}



\section{Introduction}
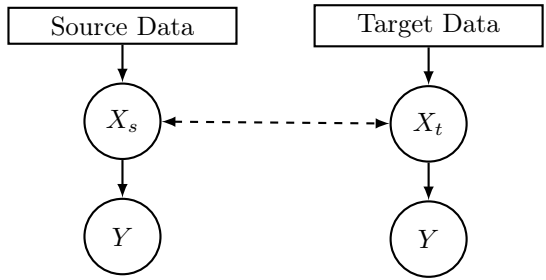
\begin{figure}[htbp]
    \centering
    \begin{tikzpicture}[->,>=latex, node distance=0.5cm, thick]

    \node[draw, rectangle, minimum width=3cm, fill=white] (Src) {Source Data};
    \node[draw, circle, below=of Src, fill=white, minimum size=1cm] (Xs) {$X_s$};
    \node[draw, circle, below=of Xs, fill=white, minimum size=1cm] (Ys) {$Y$};

    \node[draw, rectangle, minimum width=3cm, fill=white, right=1cm of Src] (Tgt) {Target Data};
    \node[draw, circle, below=of Tgt, fill=white, minimum size=1cm] (Xt) {$X_t$};
    \node[draw, circle, below=of Xt, fill=white, minimum size=1cm] (Yt) {$Y$};

    \draw[->] (Src) -- (Xs);
    \draw[->] (Xs) -- (Ys);
    \draw[->] (Tgt) -- (Xt);
    \draw[->] (Xt) -- (Yt);

    \draw[dashed, <->] (Xs) -- (Xt);

    \end{tikzpicture}
    \caption{Illustration of domain shift in medical image analysis.}
\label{fig:domain_shift}
\end{figure}

Over the past decade, significant advances in machine learning for disease diagnosis, management, and monitoring have dramatically changed clinical imaging practice. Advances in deep learning have enabled a wide range of medical image analysis tasks, including image segmentation, classification, and anomaly detection, to achieve very high statistical levels of accuracy and sensitivity across diverse tasks and imaging modalities \cite{litjens2017survey,lecun2015deep, cortes2021schizophrenia, tweneboah2025deep}. Despite these advances, several limitations remain: most medical image analysis data are unstructured, annotations are cumbersome and expensive, and model performance degrades when datasets from a particular institution or cohort are generalised to other settings. This is the generalisation problem across domains, sometimes called domain shift, which is a significant issue impairing the robustness and utility of machine learning (ML) in real-world clinical applications \cite{gulrajani2021in,subbaswamy2021evaluating,bernhardt2022investigating,wang2021harmonization,ye2021influence,zhang2021empirical,valvano2021adversarial,prosperi2020causal,hirano2021universal,huang2020causal,gordaliza2022translational,prosperi2020causal, hirano2021universal, huang2020causal, 
gordaliza2022translational}. This problem is illustrated in Figure \ref{fig:domain_shift}, where a model trained on source data ($X_s$) and corresponding labels ($Y$) may rely on spurious features or annotations that do not generalise well to target data ($X_t$) with the same labels ($Y$) \cite{zhang2022adversarial, santacruz2021metadata}.

The traditional perspective on transfer learning, which allows models trained in one context to be more efficient in others by utilising knowledge obtained elsewhere, has offered some solutions to the domain-shift problem. Transfer learning improves model accuracy with limited labelled data by pretraining on large datasets (e.g., ImageNet for general vision or domain-specific medical image datasets) and adapting those representations to smaller, task-specific settings \cite{pan2010survey, suleiman2024two}. Pretraining can use supervised learning on labelled data or self-supervised learning when labels are scarce. Foundation models trained on large, general datasets can be adapted to specific tasks, such as medical image segmentation. In contrast, task-specific pretrained models are fine-tuned directly for a particular domain. This leverages features learned from diverse datasets to enhance performance on smaller, specialised datasets \cite{pan2010survey}. However, a significant limitation of conventional transfer learning is its inability to explicitly model and leverage the underlying causal structures in the data. This limitation can lead to challenges in the medical image pipeline, as models may capture spurious relationships that do not generalise well outside the training environment, leading to degraded robustness and fairness under distribution shift \cite{pearl2009causality,scholkopf2021toward, yi2020clevrer, benkarim2021cost, budd2021survey, schrouff2022fairness, singla2021causal, ouyang2021causality, li2021knockoff, chuang2022nonlinear, ding2022carts}. For example, if a model is trained to detect pneumonia cases, it could rely on background artefacts (e.g., imaging device markers) that are not related to the presence of pneumonia, resulting in predictions that are biased and clinically unfounded \cite{subbaswamy2021evaluating, adebayo2022post}.

Causal inference is promising because it addresses these challenges by establishing and verifying cause--effect relationships within data rather than relying solely on correlations \cite{pearl2009causality, mani2000causal, polsterl2021estimation, ramsey2010six, ke2022learning, clivio2022neural, uhler2025causal, carloni2025human, mesinovic2025causal}. Unlike purely correlational patterns, causal relationships are based on intrinsic mechanisms within the data and are therefore more likely to remain stable across different environments and external contexts. In healthcare applications, for example, the causal relationship between disease biomarkers and disease presence should remain consistent across hospitals, imaging devices, and patient populations \cite{scholkopf2021toward, fehr2022transportability, fay2025mimm}.

Causal Transfer Learning (CTL) refers to the integration of causal inference principles into transfer learning frameworks, with the aim of transferring knowledge based on invariant causal relationships rather than dataset-specific correlations. By leveraging causal structures that are expected to remain stable across domains, CTL combines the adaptability of transfer learning with the robustness and interpretability of causal models. Recent advances have further demonstrated how causal representations and invariant mechanisms can support out-of-distribution generalisation and transfer across heterogeneous source domains \cite{song2026cogs,li2026transfer}.
Consequently, CTL models are expected to generalise more reliably across diverse clinical settings and unseen data distributions \cite{rojas2018invariant, zapaishchykova2021interpretable, hussain2021practical, liu2021learning}. 

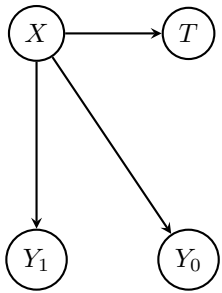
\begin{figure}[htbp]
\centering
\begin{tikzpicture}[node distance=2cm, >=stealth, thick]

\node[draw, circle] (X) {$X$};
\node[draw, circle, right of=X] (T) {$T$};
\node[draw, circle, below of=T, yshift=-1cm] (Y0) {$Y_0$};
\node[draw, circle, below of=X, yshift=-1cm] (Y1) {$Y_1$};

\draw[->] (X) -- (T);
\draw[->] (X) -- (Y1);
\draw[->] (X) -- (Y0);

\end{tikzpicture}
\caption{Ignorability assumption: potential outcomes $(Y_0, Y_1)$ are independent 
of treatment $T$ given covariates $X$.}
\label{fig:ignorability}
\end{figure}

A key assumption underlying many causal inference and CTL approaches is \emph{ignorability}, illustrated in Figure~\ref{fig:ignorability}. Ignorability assumes that, once the relevant observed covariates $X$ are accounted for, the factors influencing data acquisition or domain membership are independent of the outcome of interest. In medical image analysis, these factors may include scanner type, staining protocol, imaging device, or preprocessing pipeline, while $X$ represents observed image-derived features.

Under this assumption, any association between acquisition-related factors and clinical outcomes can be explained through the observed covariates, allowing models to adjust for confounding effects introduced by dataset-specific biases or variations in imaging conditions. This idea is central to CTL, where the objective is to distinguish spurious correlations arising from observational or domain-specific biases from stable causal relationships that generalise across domains. By explicitly accounting for confounding variables, CTL methods aim to identify invariant predictive mechanisms that remain consistent across changes in acquisition conditions, patient populations, and clinical protocols, thereby improving robustness and cross-domain generalisation \cite{fawkes2022selection, rubin1978bayesian, diaz2026invariance}.

Recent surveys have highlighted the growing importance of causality in medical image analysis, showing how frameworks such as structural causal models, counterfactual reasoning, do-calculus, potential outcomes, and graphical models can improve interpretability, robustness, and diagnostic precision \cite{vlontzos2022review, neuberg2003causality, boge2025causality, austin2011introduction, sedgewick2019mixed}. However, these works primarily examine causality within medical imaging itself rather than how causal reasoning can be systematically integrated with cross-domain transfer. Closely related recent surveys address adjacent problems from different perspectives. Atasever et al. review transfer learning in medical image analysis broadly, focusing on methods, datasets, trends, and limitations, but without organising the field around causal assumptions or causal invariance \cite{atasever2023comprehensive}. Yoon et al. review domain generalisation for medical image analysis and categorise methods into data-, feature-, model-, and analysis-level approaches across the imaging workflow, providing an important account of out-of-distribution robustness, yet without centring causal inference as the transfer principle or synthesising methods through structural causal models, interventions, and counterfactual reasoning \cite{yoon2024domain}. Matta et al. provide a systematic review of generalisation research in medical image classification, proposing a shift-type taxonomy based on 77 studies, but their scope is restricted to classification and does not unify causal and transfer-learning perspectives across imaging tasks \cite{matta2024systematic}. Taken together, these surveys establish the relevance of causality, transfer, and generalisation in medical imaging, but they leave open the specific problem addressed here: a unified synthesis of causal transfer learning for medical image analysis, where domain shift is treated explicitly as a causal problem. Building on these prior surveys, this review focuses specifically on CTL in medical image analysis, namely methods that combine causal reasoning with transfer, adaptation, or generalisation under domain shift. Rather than offering another broad review of transfer learning, domain generalisation, or classification-focused robustness, we synthesise the intersection of these areas and organise the literature through a unified framework that links causal formalisms (e.g., structural causal models and potential outcomes), causal operations (e.g., invariance, back-door adjustment, and counterfactual generation), transfer settings (e.g., domain adaptation, domain generalisation, source-free transfer, and multimodal transfer), imaging tasks, and shift types. As summarised in Table~\ref{tab:survey_comparison}, this integrated perspective makes the distinctive contribution of the present review explicit: it links causal framework, causal operation, transfer setting, imaging task, and shift type within a single synthesis, thereby clarifying where causal assumptions meaningfully improve cross-domain robustness, fairness, and clinical reliability, and where current claims remain methodologically or empirically underdeveloped.

\begin{table*}[h]
    \centering
    \scriptsize
    \caption{Comparison between existing surveys and this work.}
    \label{tab:survey_comparison}
    \begin{tabular}{@{}>{\raggedright\arraybackslash}p{2.6cm}
                    >{\raggedright\arraybackslash}p{2.5cm}
                    >{\raggedright\arraybackslash}p{4.1cm}
                    >{\raggedright\arraybackslash}p{4.1cm}@{}}
        \toprule
        \textbf{Survey} & \textbf{Primary focus} & \textbf{What it covers well} & \textbf{Gap relative to this survey} \\
        \midrule

        Atasever et al. \cite{atasever2023comprehensive}
        & Transfer learning in medical image analysis
        & Broad review of methods, datasets, trends, knowledge gaps, constraints, and limitations in medical-image transfer learning
        & Does not organise the field around causal assumptions, invariance, interventions, or counterfactual reasoning \\
        \midrule

        Yoon et al. \cite{yoon2024domain}
        & Domain generalisation in medical image analysis
        & Structured review of out-of-distribution robustness across data-, feature-, model-, and analysis-level DG methods throughout the imaging workflow
        & Does not treat domain shift as a causal problem or synthesise methods through SCMs, potential outcomes, interventions, or counterfactual transfer \\
        \midrule

        Matta et al. \cite{matta2024systematic}
        & Generalisation in medical image classification
        & Systematic review of classification studies under distribution shift, with a taxonomy organised by shift type
        & Restricted to classification and does not unify causal and transfer-learning perspectives across imaging tasks \\
        \midrule

        Vlontzos et al. \cite{vlontzos2022review}
        & Causality in medical image analysis
        & Reviews causal reasoning methods for medical imaging AI/ML, highlighting their relevance to robustness and clinical translation
        & Focuses on causality in imaging broadly rather than on transfer, adaptation, or generalisation under domain shift as the central organising problem \\
        \midrule

        Methodological causal foundations \cite{neuberg2003causality,boge2025causality,austin2011introduction,sedgewick2019mixed}
        & General causal inference frameworks
        & Establish core tools such as do-calculus, propensity-score reasoning, and graphical causal models that underpin later CTL methods
        & Provide conceptual foundations rather than a survey of causal transfer learning, domain shift, or medical-image-analysis transfer methods \\
        \midrule

        \textbf{This survey}
        & \textbf{Causal transfer learning in medical image analysis}
        & \textbf{Unified synthesis of methods combining causal reasoning with transfer, adaptation, and generalisation under domain shift}
        & \textbf{Organises the field jointly by causal framework, causal operation, transfer setting, imaging task, and shift type} \\
        \bottomrule
    \end{tabular}

    \vspace{1mm}
    \parbox{\textwidth}{\footnotesize \textit{Note:} As represented by the surveys included in this table, closely related review spaces include transfer learning, domain generalisation, classification-focused generalisation, and causality in medical imaging. The present review focuses specifically on their intersection in causal transfer learning under domain shift.}
\end{table*}

\begin{figure}[htbp]
    \centering
    \includegraphics[width=0.9\linewidth]{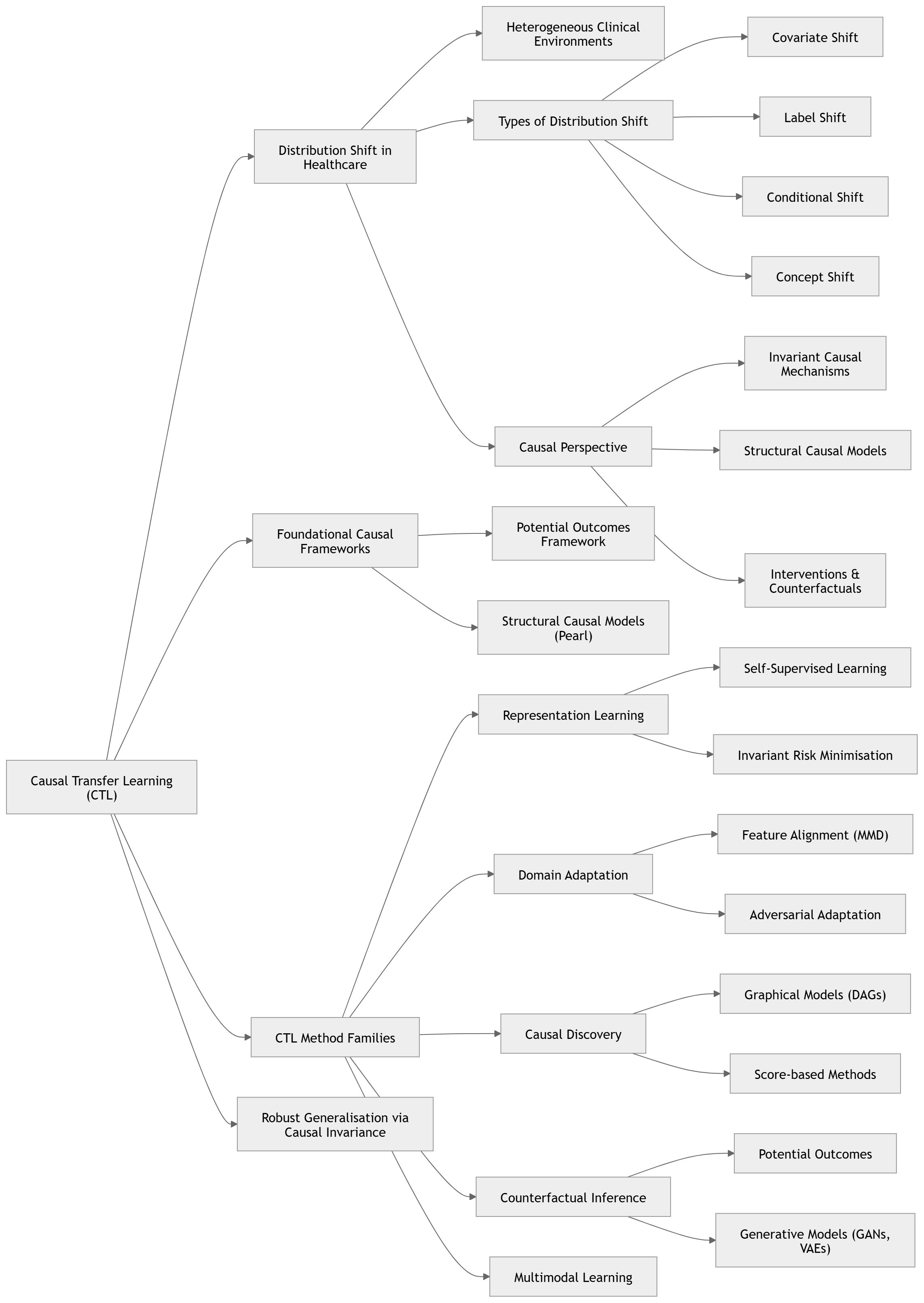}
\caption{Taxonomy of CTL methods organised across (i) foundational causal frameworks (Structural Causal Models and Potential Outcomes), (ii) core methodological strategies including domain adaptation, causal discovery, invariant representation learning, and counterfactual inference, (iii) supporting causal structures such as ignorability and multimodal environment modelling, and (iv) task-oriented deployment in medical image analysis. Each methodological family is further interpreted in relation to the four types of distribution shift (covariate, label, conditional, and concept shift), highlighting how CTL methods operate at different levels of the data-generating process.}
\label{fig:ctl-taxonomy}
\end{figure}


Unlike conventional surveys that separately review causal inference, transfer learning, domain adaptation, or self-supervised learning, this survey organises the literature around the problem of distribution shift in medical imaging. Specifically, we analyse how different forms of shift (including covariate shift, label shift, conditional shift, and concept shift) affect the generalisability of medical AI systems, and we examine which learning paradigms can robustly address each type. Within this framework, conventional transfer learning and domain adaptation methods are discussed in terms of their ability to handle statistical distribution mismatches. In contrast, causal transfer learning is presented as a principled approach to learning invariant mechanisms across changing environments. This distribution-shift-centric perspective provides a more unified understanding of robustness, generalisation, and causal invariance in healthcare AI.

Despite recent progress, the surveyed literature reveals several unresolved theoretical, methodological, and clinical challenges in causal transfer learning for medical image analysis. First, it remains unclear under what conditions invariant causal mechanisms can be reliably identified across heterogeneous clinical environments, particularly when imaging protocols, annotation practices, and disease distributions evolve simultaneously. Second, although many CTL methods assume that learned latent representations correspond to causal factors, the extent to which current deep representation learning frameworks truly recover causal structure rather than stable correlations remains largely unverified in high-dimensional medical imaging settings. Third, existing evaluation protocols continue to rely predominantly on predictive metrics such as Dice score, AUC, and classification accuracy, which do not directly assess causal validity, counterfactual consistency, or robustness under intervention and distribution shift. Moreover, an important unresolved tension exists between enforcing domain invariance and preserving clinically meaningful population-specific information, as overly aggressive invariance constraints may suppress diagnostically relevant variations across demographic groups or imaging protocols. Finally, integrating causal reasoning into real-world clinical systems poses challenges related to uncertain causal graphs, latent confounding, and dynamically evolving medical knowledge. Addressing these open challenges will be essential for developing clinically reliable, interpretable, and genuinely generalisable CTL frameworks.

The main contributions of this survey are:

\begin{itemize}

\item We reinterpret distribution shift in healthcare (including environmental, institutional, and population-induced variability) as perturbations of underlying causal mechanisms, rather than purely statistical discrepancies. This reframes domain adaptation, domain generalisation, and multi-institutional learning as problems of learning invariant causal structure across environments.

\item We introduce a new taxonomy (see Figure \ref{fig:ctl-taxonomy}) that organises causal transfer learning methods according to (i) their causal framework (structural causal models or potential outcomes), (ii) the type of causal operation they implement (invariance learning or counterfactual reasoning), and (iii) their role in the transfer pipeline (representation learning, alignment, or decision-making).

\item We provide a comprehensive and structured review of more than 82 studies on causal transfer learning for medical imaging under distribution shift, covering classification, segmentation, reconstruction, enhancement, and anomaly detection tasks across diverse shift scenarios, including scanner, protocol, population, pathology, and modality variations.

\item We summarise publicly available datasets, benchmarks, and reported cross-domain performance gains, providing evidence on when causal assumptions are empirically supported versus when they remain theoretical.

\item We analyse how causal transfer learning improves robustness, fairness, and trustworthiness in multimodal, longitudinal, and federated clinical settings by explicitly accounting for invariant and non-invariant data-generating mechanisms.

\item Beyond surveying existing methodologies, this review also highlights unresolved theoretical and clinical challenges in CTL, including questions surrounding causal identifiability, representation validity, evaluation under intervention and distribution shift, and the trade-off between invariance and clinically meaningful variability.

\end{itemize}

A structured literature review was conducted across PubMed, IEEE Xplore, Scopus, Web of Science, ACM Digital Library, ScienceDirect, SpringerLink, and Google Scholar to identify studies published between 2010 and 2026. Search queries combined terms related to causal inference, causal learning, transfer learning, domain adaptation, domain generalisation, distribution shift, and medical imaging. After deduplication, studies were screened using predefined inclusion and exclusion criteria. Eligible studies were required to address medical image analysis while integrating causal reasoning with transfer learning, domain adaptation, or domain generalisation under distribution shift, and to report methodological or quantitative experimental results. Excluded studies included duplicate records, non-medical or non-imaging applications, purely conceptual works without imaging implementation, conventional transfer learning approaches lacking causal reasoning, review articles, and papers with insufficient technical detail for comparative analysis. Following title/abstract screening and full-text assessment, more than 82 studies were retained for analysis. The final corpus was organised according to imaging task (classification, segmentation, reconstruction, enhancement, and anomaly detection), distribution shift type (scanner, protocol, population, pathology, and modality), and causal assumptions such as invariance, intervention, and confounding control. This organisation enabled a systematic comparison of how causal transfer learning improves robustness and generalisation relative to correlation-based approaches across diverse shift settings.
s

The structure of this paper is organised as follows: Section 2 discusses the concepts of distribution Shift in healthcare and motivates the need for CTL. Section 3 presents an overview of CTL and its integration of causal reasoning into cross-domain knowledge transfer to achieve robust and invariant representations across heterogeneous datasets. Section 4 details the core methodologies underpinning CTL, including domain adaptation, causal discovery, and counterfactual inference, and highlights their distinct roles and recent advances. Section 5 reviews key applications of CTL in medical image analysis and discusses their methodological implications. Section 6 provides an in-depth review of CTL applied specifically to medical image analysis, synthesising state-of-the-art techniques, datasets, results, and clinical relevance. Section 7 examines the main challenges and limitations that hinder the adoption of CTL in clinical practice. Section 8 outlines current research directions to address these limitations and expand CTL’s impact. Section 10 discusses the potential of CTL to address domain shifts and data scarcity in medical image analysis, emphasising the need for diverse, large datasets and highlighting those well-suited to CTL. Section 9 evaluates causal learning models, contrasting causal-specific assessment criteria with traditional image analysis metrics that prioritise correlation-based performance. Section 10 discusses broader developments beyond causal AI, situating CTL within emerging paradigms of adaptive, multimodal, and decision-centric intelligence. Finally, Section 11 concludes by emphasising the transformative potential of CTL in medical image analysis and its ability to advance reliable, interpretable, and generalisable clinical AI systems.


\section{Causal Inference and Transfer Learning under Distribution Shift in Healthcare}

Healthcare machine learning operates across heterogeneous clinical environments, including hospitals, populations, imaging devices, and clinical protocols. These variations induce distribution shifts that violate the i.i.d. assumption and degrade model performance when deployed outside the training domain. Let $X$ denote observed patient data (e.g., imaging features, clinical measurements, or multimodal inputs) and $Y$ denote clinical outcomes (e.g., diagnosis or prognosis), both generated within an environment indexed by $e$.

\begin{equation}
P_e(X,Y) \neq P_{e'}(X,Y)
\end{equation}

where $e$ and $e'$ denote different clinical environments. From a causal perspective, although observational distributions vary across environments, the underlying data-generating mechanisms governing $(X,Y)$ may remain invariant. The goal of causal transfer learning is therefore to identify and exploit these invariant structures to ensure robust generalisation across environments \cite{pearl2009causality, yao2021survey}.

\subsection{Distribution Shifts in Healthcare}

Distribution shifts can be systematically characterised by changes in components of the factorisation $P(X,Y)=P(Y|X)P(X)$. These shifts determine the extent to which transfer learning, domain adaptation, and causal inference are required.

\subsubsection{Covariate Shift}

\begin{equation}
P_s(X) \neq P_t(X), \quad P_s(Y|X)=P_t(Y|X)
\end{equation}

Covariate shift occurs when the input distribution changes across environments while the predictive mechanism remains stable. This is common in medical imaging due to differences in scanners, acquisition protocols, and patient populations \cite{pan2010survey, kouw2019smoothness}.

In this setting, transfer learning is often effective because pretrained representations remain useful when low-level structure is shared across domains. This is further strengthened by self-supervised learning (SSL), which learns representations without labels:

\begin{equation}
L = \mathbb{E}_{(x,y)\sim D}[\ell(f(x), g(y))]
\end{equation}

where $y$ may represent augmented views or proxy targets depending on the architecture \cite{chen2020simple, radford2021learning}. Such representations reduce sensitivity to shifts in $P(X)$ by learning invariant feature encodings.

\subsubsection{Label Shift}

\begin{equation}
P_s(Y) \neq P_t(Y), \quad P_s(X|Y)=P_t(X|Y)
\end{equation}

Label shift arises when disease prevalence or population composition differs across environments, such as when incidence rates vary across hospitals or regions. Although the class-conditional distribution remains stable, the marginal distribution over outcomes changes.

In this case, transfer learning alone is insufficient. Correction typically requires reweighting or calibration of class priors, often within a Bayesian framework that explicitly models uncertainty in label distributions \cite{gelman2013bayesian}.

\subsubsection{Conditional Shift}

\begin{equation}
P_s(X|Y) \neq P_t(X|Y)
\end{equation}

Conditional shift reflects changes in how diseases manifest across populations, imaging devices, or clinical protocols. For example, tumour morphology or radiological appearance may differ across scanners or demographic groups \cite{castro2020causality, petersen2023demographically}.

Classical domain adaptation methods attempt to mitigate this by aligning feature distributions. A widely used approach is Maximum Mean Discrepancy (MMD):

\begin{equation}
\text{MMD}(P_s, P_t)=
\left\|
\frac{1}{n_s}\sum_{i=1}^{n_s}\phi(x_i^s) -
\frac{1}{n_t}\sum_{j=1}^{n_t}\phi(x_j^t)
\right\|_{\mathcal{H}}^2
\end{equation}

where $\phi$ maps data into a reproducing kernel Hilbert space (RKHS) \cite{gretton2012kernel}. However, purely statistical alignment may fail when the conditional data-generating process changes, motivating causal representation learning that separates invariant disease mechanisms from environment-specific variations.

\subsubsection{Concept Shift}

\begin{equation}
P_s(Y|X) \neq P_t(Y|X)
\end{equation}

Concept shift is the most severe form of distribution shift, in which the mapping between features and outcomes changes due to shifts in diagnostic criteria, clinical guidelines, or outcome definitions. In this regime, standard transfer learning fails because the predictive function itself is not invariant.




\subsection{From Distribution Shift to Causal Transfer Learning}

Across the different types of distribution shift, a key observation is that their difficulty increases as one moves from changes in marginal distributions to changes in the predictive mechanism itself.

In particular, covariate shift ($P(X)$ changes), label shift ($P(Y)$ changes), and conditional shift ($P(X|Y)$ changes) can often be partially addressed using representation learning, reweighting strategies, or domain alignment techniques. These approaches rely on the assumption that some component of the data-generating process remains stable across environments.

However, this assumption breaks down under concept shift, where $P(Y|X)$ changes. In this case, the learned mapping between features and outcomes is no longer invariant, meaning that neither standard transfer learning nor conventional domain adaptation can guarantee generalisation across environments.

This breakdown highlights the need for modelling approaches that go beyond statistical alignment and instead focus on structural invariance in the data-generating process. This perspective naturally leads to causal transfer learning, which is introduced in the following section.

\section{Causal Transfer Learning}

CTL emerges as a unifying framework for addressing the limitations of conventional transfer learning under distribution shift, particularly in settings where the predictive relationship between inputs and outcomes is not stable across environments. Unlike standard approaches that primarily focus on aligning marginal distributions or feature spaces, CTL explicitly constrains adaptation to respect the underlying causal structure of the data-generating process.

Rather than relying solely on statistical similarity between domains, CTL focuses on identifying, preserving, or reconstructing causal mechanisms that remain invariant across environments, while explicitly accounting for those components that change under shift. As highlighted in previous sections, failures of standard transfer learning are most pronounced under conditional and concept shifts, where changes in $P(X|Y)$ or $P(Y|X)$ break the assumption of stable correlations. CTL addresses this limitation by targeting invariant causal mechanisms that persist across changes in acquisition conditions, patient populations, imaging devices, and clinical protocols, thereby enabling robust cross-domain generalisation.

Within this survey, CTL is defined as the integration of causal inference principles into transfer learning frameworks to improve robustness under distribution shift (Table \ref{Tablekeyconcepts}). Rather than learning correlations that may be environment-specific, CTL methods aim to identify representations that correspond to stable causal relationships, thereby improving generalisation across heterogeneous datasets and clinical settings.

Here, we define causal transfer learning as the integration of causal inference into transfer learning to improve robustness and efficiency when training models across domains (see Table \ref{Tablekeyconcepts}). The result is a model representation that is invariant across different datasets while modelling the underlying causal relationships among variables. This is important in practice because models that train well on one dataset often need to generalise well to another dataset, possibly under quite different conditions, which is particularly relevant in medical image analysis. 

Mathematically, we can express causal transfer learning as:

\begin{equation}
    \min_{\theta} \mathbb{E}_{(X_{\text{causal}},Y) \sim D_t} [\mathcal{L}(f(X_{\text{causal}}; \theta), Y)] 
    + \lambda \, \mathbb{E}_{(X_{\text{causal}},Y) \sim D_s} [\mathcal{L}(f(X_{\text{causal}}; \theta), Y)],
\end{equation}

where $\mathcal{L}$ denotes a task-specific loss function, $f$ is a predictive model parameterised by $\theta$, $D_s$ and $D_t$ denote source and target domains respectively, and $\lambda$ balances their contributions \cite{rojas2018invariant, teshima2020few}. In CTL, the learned representations are assumed to correspond to causal mechanisms that remain invariant across domains, enabling transfer beyond purely statistical alignment.

\begin{table*}[h]
    \centering
    \scriptsize
    \caption{Key Concepts in Causal Transfer Learning}
    \begin{tabular}{@{}>{\raggedright}p{3cm} p{5cm} p{5cm}@{}}
        \toprule
        \textbf{Concept} & \textbf{Definition} & \textbf{Importance in CTL} \\
        \midrule
        Causal Inference & The process of deducing causal relationships from data, often through techniques such as DAGs \cite{pearl2009causality}. & Provides a basis for understanding how different features relate to outcomes, leading to better model generalisation. \\
        \midrule
        Transfer Learning & A machine learning approach where a model trained on one task is adapted for another related task \cite{pan2010survey}. & Enables leveraging knowledge from source domains to improve performance in target domains. \\
        \midrule
        Domain Adaptation & Techniques aimed at minimising differences between source and target domains, thereby improving model performance \cite{kouw2019review}. & Addresses shifts between patient populations and imaging conditions. \\
        \midrule
        Counterfactual Inference & A framework for exploring hypothetical scenarios to evaluate the impact of interventions, often using potential outcomes \cite{rubin1974estimating}. & Provides insights into causal relationships, aiding personalised treatment reasoning. \\
        \midrule
        Invariant Risk Minimisation & A method that focuses on learning features that remain consistent across different environments to enhance model transferability \cite{arjovsky2019invariant}. & Addresses domain shift by encouraging stable predictive mechanisms across environments. \\
        \bottomrule
    \end{tabular}
\label{Tablekeyconcepts}
\end{table*}

\subsection{Invariant Risk Minimisation}

Invariant Risk Minimisation (IRM) is a specific strategy within causal transfer learning that aims to identify model representations that remain stable across multiple environments or domains. The idea is that by focusing on invariant features, the model can better generalise its predictions to new or unseen data.

Formally, given multiple environments \( E_1, E_2, \ldots, E_k \), the IRM objective is defined as:

\begin{equation}
    \min_{\theta} \sum_{i=1}^k \mathbb{E}_{(X,Y) \sim D_i} \left[ \mathcal{L}(f(X; \theta), Y) \right] + \eta \cdot \mathbb{E}_{(X,Y) \sim D_i} \left[ \left( \nabla_{\theta} \mathcal{L}(f(X; \theta), Y) \right)^2 \right],
\end{equation}

where \( \eta \) is a hyperparameter that controls the trade-off between the primary objective of minimising risk and the secondary objective of ensuring invariance across environments \cite{arjovsky2019invariant}. The first term in this equation computes the average loss across all environments, while the second term penalises the model based on the variance of gradients across environments. By minimising this objective, IRM encourages the model to focus on features that contribute to consistent performance across environments.

From a causal perspective, IRM can be interpreted as encouraging invariance of predictive mechanisms across environments, although it does not explicitly model causal structure. As such, it can be seen as a partial step towards causal representation learning within CTL frameworks.

Through these methodologies, causal transfer learning and IRM not only enhance model adaptability across domains but also deepen understanding of causal relationships in the data.

\subsection{Counterfactual Inference}

Counterfactual inference plays a vital role in causal transfer learning by enabling the evaluation of hypothetical scenarios, thereby deepening understanding of causal effects. This process can be implemented by utilising potential outcomes or by employing generative models to simulate counterfactual scenarios \cite{balke1994probabilistic, reynaud2022dartagnan} (see Table \ref{TableCIT}). The causal effect can be estimated as follows:

\begin{equation}
    \text{ATE} = \mathbb{E}[Y(1)] - \mathbb{E}[Y(0)],
\end{equation}

where ATE refers to the Average Treatment Effect \cite{imbens2015causal}. Understanding these counterfactuals is crucial for assessing how different interventions or conditions may yield varying diagnostic/prognostic outcomes, particularly in medical image analysis.

\begin{table*}[h]
    \centering
    \scriptsize
    \caption{Counterfactual Inference Techniques}
    \begin{tabular}{@{}>{\raggedright}p{4cm} p{5cm} p{5cm}@{}}
        \toprule
        \textbf{Technique} & \textbf{Description} & \textbf{Application in medical image analysis} \\
        \midrule
        Potential Outcomes Framework & A framework for evaluating treatment effects based on hypothetical scenarios \cite{imbens2015causal}. & Estimating the effect of different treatment plans on patient outcomes in clinical trials. \\
        \midrule
        Counterfactual Generative Models & Models that generate synthetic data based on counterfactual scenarios, facilitating the exploration of treatment effects \cite{battaglia2018relational}. & Simulating patient responses under different treatment conditions to inform decision-making. \\
        \midrule
        Causal effect estimation and quantification of the impact of interventions on outcomes using observational data, often through techniques such as propensity score matching \cite{rosenbaum1983central}. & Estimating the effect of imaging modality on diagnostic accuracy across different patient groups. \\
        \bottomrule
    \end{tabular}
    \label{TableCIT}
\end{table*}

\subsubsection{Generative Counterfactual Models}

Generative models serve as powerful tools for simulating counterfactual outcomes, leveraging methods such as Generative Adversarial Networks (GANs) \cite{vlontzos2021twin, goodfellow2014generative}. The GAN framework can be represented mathematically as:

\begin{equation}
    \min_{G} \max_{D} \mathbb{E}_{x \sim P_{data}}[\log D(x)] + \mathbb{E}_{z \sim P_z}[\log(1 - D(G(z)))]],
\end{equation}

where \( G \) denotes the generator and \( D \) signifies the discriminator. In a counterfactual context, \( G \) generates data representing potential outcomes given a specific intervention or treatment scenario \cite{goodfellow2014generative}. This approach is particularly valuable in medical image analysis, as it enables exploration of how diagnostic results may change across different imaging techniques or patient conditions.

Within CTL, such generative counterfactual models provide a mechanism to simulate interventions, enabling evaluation of how changes in imaging conditions or clinical decisions affect downstream predictions.

\section{Causal Transfer Learning Strategies}

Within the CTL framework, methodological strategies can be organised according to the type of distribution shift they are most closely associated with, as well as the causal mechanism they aim to preserve or recover. In this view, domain adaptation primarily addresses marginal shifts, causal discovery targets structural (conditional) shifts, and counterfactual inference becomes essential when the predictive mechanism itself is no longer stable under concept shift. Representation learning approaches, such as invariant risk minimisation, act as a bridging strategy that promotes stability across multiple shift regimes.

More generally, these strategies operate over different components of the data-generating process, ranging from marginal distributions to structural causal relationships. This provides a principled way to understand when each method is expected to succeed or fail under heterogeneous clinical environments.

\subsection{Domain Adaptation in Causal Transfer Learning}

Domain adaptation (DA) is vital in CTL to mitigate domain-shift effects on model performance. DA techniques are categorised into three primary strategies: instance reweighting, feature alignment, and adversarial training, along with emerging approaches that enhance adaptability (see Table \ref{TableDA}).

\begin{table*}[h]
    \centering
    \scriptsize
    \caption{Domain Adaptation Techniques in Causal Transfer Learning}
    \begin{tabular}{@{}>{\raggedright}p{4cm} p{5cm} p{5cm}@{}}
        \toprule
        \textbf{Technique} & \textbf{Description} & \textbf{Example Application in medical image analysis} \\
        \midrule
        Instance Reweighting & Assigning weights to training samples based on their relevance to the target domain \cite{xia2018instance, guan2021domain}. & Adjusting weights of training samples in a chest X-ray dataset to reflect the demographics of a local population. \\
        \midrule
        Feature Alignment & Transforming feature representations to minimise differences between source and target domains using techniques such as MMD \cite{gretton2012kernel}. Unlike distillation, which transfers knowledge between models, feature alignment focuses on aligning feature distributions across domains. & Aligning image feature distributions across different imaging devices, such as MRI scanners. \\
        \midrule
        Adversarial Training & A method where a model is trained to confuse a domain classifier, forcing learned features to be domain-invariant \cite{goodfellow2014generative}. & Using adversarial networks to improve robustness in skin lesion classification across various imaging protocols. \\
        \bottomrule
    \end{tabular}
    \label{TableDA}
\end{table*}

\begin{enumerate}

\item \textbf{Instance Reweighting.} This strategy involves assigning weights to training samples in the source domain based on their relevance to the target domain. The weight \( w_i \) for each instance \( x_i \) can be calculated as:
\begin{equation}
w_i = \frac{p_t(x_i)}{p_s(x_i)}.
\end{equation}

This approach is primarily effective under covariate shift, where differences between domains arise from changes in $P(X)$ while the conditional mechanism $P(Y|X)$ remains approximately invariant. It may also partially mitigate label shift when reweighting can correct changes in $P(Y)$.

\item \textbf{Feature Alignment.} This technique adjusts feature spaces to align the distributions of the source and target domains using metrics such as MMD:
\begin{equation}
\text{MMD}(P, Q) =
\left\|
\frac{1}{n_1} \sum_{i=1}^{n_1} \phi(x_{1,i}) -
\frac{1}{n_2} \sum_{j=1}^{n_2} \phi(x_{2,j})
\right\|_{\mathcal{H}}^2,
\end{equation}

where \( \phi \) maps data into a reproducing kernel Hilbert space \cite{gretton2012kernel}.

Feature alignment methods extend beyond pure covariate shift and can improve robustness under mild conditional shift. However, they do not explicitly guarantee preservation of causal mechanisms when structural relationships such as $P(X|Y)$ vary across environments.

\item \textbf{Adversarial Training.} Inspired by GANs, adversarial domain adaptation employs a dual network architecture:
\begin{equation}
\mathcal{L}_{domain} =
-E_{x \sim P}[\log(D(x))] -
E_{x \sim Q}[\log(1 - D(x))].
\end{equation}

This encourages learning of domain-invariant representations by removing domain-specific signals from feature embeddings. While effective under covariate shift, it may fail under concept shift, where the predictive mechanism $P(Y|X)$ itself changes across environments.

\end{enumerate}

Overall, domain adaptation methods operate primarily at the level of marginal distribution alignment and are therefore limited when shifts affect the underlying causal structure rather than surface-level statistics.

\subsection{Causal Discovery Techniques}

This step involves identifying causal variables and their relationships within medical image analysis data, which is critical for CTL. A subset of the principal causal discovery methods relevant to this setting is summarised in Table~\ref{TableCDM}.

Unlike domain adaptation, which focuses on distribution alignment, causal discovery aims to recover invariant structural relationships that remain stable across environments. This makes it particularly relevant under conditional shift, where $P(X|Y)$ may vary but the underlying causal graph remains consistent.

\textbf{1. Graphical Models.}
Directed Acyclic Graphs (DAGs) are commonly used to represent causal relationships \cite{vowels2021dya}. Methods include constraint-based and score-based approaches.

Score-based methods optimise:
\begin{equation}
\text{BIC} = \log(L) - \frac{k}{2} \log(n),
\end{equation}

where $L$ is the likelihood, $k$ the number of parameters, and $n$ the sample size. These approaches aim to recover causal structure that is invariant across environments, enabling more stable transfer under heterogeneous clinical conditions.

\textbf{2. Function-based Causal Discovery.}
These methods model functional relationships such as ANMs and LiNGAM.

These approaches assume that the underlying generative mechanisms remain invariant even when observed conditional distributions change, making them particularly useful for handling structured shifts in medical imaging data.

\textbf{3. Gradient-based Causal Discovery.}
Methods such as NOTEARS and GOLEM learn causal structure via optimisation.

These scalable approaches are especially relevant in high-dimensional settings such as medical imaging, where multiple overlapping forms of distribution shift may occur simultaneously.

\textbf{4. Invariant Representation Learning (IRM).}

Invariant Risk Minimisation (IRM) can be viewed as a representation learning strategy within CTL that seeks to identify features whose predictive relationship with outcomes remains stable across environments.

\begin{equation}
\min_{\theta} \sum_{i=1}^k \mathbb{E}_{(X,Y) \sim D_i} \left[ \mathcal{L}(f(X; \theta), Y) \right] + \eta \cdot \mathbb{E}_{(X,Y) \sim D_i} \left[ \left( \nabla_{\theta} \mathcal{L}(f(X; \theta), Y) \right)^2 \right].
\end{equation}

Although IRM does not explicitly model causal structure, it encourages invariance across environments and can be interpreted as a bridge between statistical learning and causal representation learning within CTL.

\textbf{5. Contextual and Environment-Aware Learning.}

Contextual learning introduces explicit modelling of environment variables such as hospital site, scanner type, or population subgroup. These variables act as indicators of distribution shift and are particularly relevant for capturing label and conditional shifts in heterogeneous clinical settings.

\subsection{Advanced Counterfactual Inference}

Counterfactual inference is an essential component of CTL, enabling reasoning about hypothetical interventions and unobserved outcomes.

Unlike domain adaptation and causal discovery, counterfactual inference becomes most critical under concept shift, where the mapping $P(Y|X)$ itself is no longer stable across environments.

\textbf{1. Potential Outcomes Framework.}
\begin{equation}
\text{Causal Effect} = E[Y(1)] - E[Y(0)].
\end{equation}

This formulation enables estimation of treatment effects under alternative interventions and is particularly relevant in clinical settings where decision rules or outcome definitions vary across institutions.

\textbf{2. Counterfactual Generative Models.}
GANs and VAEs generate alternative outcomes \cite{kingma2019introduction, doersch2016tutorial}.

These models enable simulation of unobserved outcomes under alternative interventions, providing a mechanism for reasoning under concept shift where direct generalisation fails.

\textbf{3. Dynamic Causal Effect Estimation.}
\begin{equation}
e(x) = P(T=1 | X=x).
\end{equation}

These methods remain effective under covariate and label shift, but require integration with structural causal models to ensure validity under more severe structural shifts.

\subsection{Supporting Causal Structures in CTL}

Beyond individual methodological classes, CTL relies on several supporting causal assumptions and structures that enable valid transfer across environments. These include assumptions such as ignorability, exchangeability, and stable unit treatment effects, which underpin the identifiability of causal effects. In addition, multimodal learning settings introduce structured dependencies across heterogeneous data sources, which can be interpreted through a causal factorisation perspective. Together, these elements provide the foundational conditions under which CTL methods can reliably generalise across clinical environments.

\begin{table*}[h]
    \centering
    \scriptsize
    \caption{Causal Discovery Methods}
    \begin{tabular}{@{}>{\raggedright}p{4cm} p{5cm} p{5cm}@{}}
        \toprule
        \textbf{Method} & \textbf{Description} & \textbf{Strengths and Limitations} \\
        \midrule
        Graphical Models & Utilised DAGs to represent causal relationships and identify dependencies \cite{pearl2000causality}. & Provides a clear visual representation of causal relationships; sensitive to model assumptions. \\
        \midrule
        Invariant Risk Minimisation & Focuses on learning features that are invariant across environments to enhance generalisability \cite{arjovsky2019invariant}. & Promotes robustness to domain shifts; implementation can be complex and data-intensive. \\
        \midrule
        Causal Bayesian Networks & Combines prior knowledge and observational data to infer causal relationships and dependencies \cite{heckerman1998tutorial, borboudakis2014scoring, constantinou2023impact}. & Incorporates expert knowledge; computationally intensive and requires large datasets. \\
        \bottomrule
    \end{tabular}
\label{TableCDM}
\end{table*}


%
\section{Applications}

From the perspective of causal transfer learning, the following applications differ not only by task type, but also by the type of distribution shift they primarily address. In classification, segmentation, anomaly detection, and multimodal learning, CTL is most often used to mitigate covariate shift and conditional shift caused by differences in scanners, acquisition protocols, patient populations, and disease appearance across environments. In predictive modelling, longitudinal analysis, and some radiogenomic settings, CTL can additionally help address label shift and, more importantly, concept shift, where disease prevalence, treatment-response mechanisms, or diagnostic criteria change across settings. Because concept shift corresponds to a change in the predictive mechanism itself, it is the most difficult form of shift and is the setting in which causal transfer learning is especially important.

In the following subsections, we discuss each application in terms of both its clinical role and the dominant type(s) of distribution shift it addresses, with particular attention to concept shift as the most difficult setting, where causal transfer learning is most clearly distinguished from purely correlational transfer methods.








\subsection{Image Classification}

The primary application domain of CTL is image classification, which involves categorising medical images by content. CTL enhances the robustness of classification models, particularly in scenarios involving heterogeneous datasets \cite{zhang2024mixed}. These datasets often originate from different patient populations or imaging protocols, which can introduce significant variability in medical image features. This application most commonly addresses covariate shift, arising from changes in scanners, acquisition protocols, or patient populations, and conditional shift, when the visual manifestation of disease differs across environments. In some clinically evolving settings, classification may also face concept shift, for example when diagnostic criteria or labelling practices change over time; such cases are particularly difficult because the mapping from image features to labels is no longer stable and therefore require causal modelling rather than statistical transfer alone.

For instance, in the classification of chest X-rays, a CTL would utilise causal inference methods in order to model how given image features explicitly relate to specific disease categories, such as patterns of opacity associated with types of pneumonia \cite{rajpurkar2017chexnet, rasal2025causal}. By incorporating graphical models of these relationships, we can seamlessly integrate clinical metadata, such as age and medical history, to enhance predictive performance.

Mathematically, this can be expressed as:
\begin{equation}
    Y = f(X, C) + \epsilon
\end{equation}

where \(Y\) is the disease category, \(X\) represents the imaging features, \(C\) is the clinical metadata, and \(\epsilon\) is the error term. This enables the model to capture complex interactions and improve classification accuracy.

\subsection{Segmentation Tasks}

Image Segmentation is the process of identifying and delineating specific anatomical structures from images, such as tumours or organs. CTL strengthens segmentation by infusing it with the causal relationships of image features with biological structures-a critical step toward delineation. Segmentation tasks primarily address covariate shift, caused by differences in scanners, imaging protocols, and site-specific acquisition characteristics, and conditional shift, when anatomical or pathological appearance varies across populations or modalities. True concept shift is less common in segmentation, but may arise when annotation protocols or clinical definitions of target structures change across institutions, in which case causal modelling can help preserve anatomically meaningful invariances.

\subsubsection{Semantic Segmentation}


In the semantic segmentation of anatomical structures and pathological regions, CTL can improve the identification of these regions. This is the case when, on MRI scans, CTL allows identifying tissues as healthy or diseased through its modelling of the causality imposed by image features on semantic categories \cite{ouyang2022causality, chen2025generalizable, liang2025multimodal}.

Using a CTL framework, we can now use deep learning models such as U-Net \cite{ronneberger2015u}, with causality embedded in the loss function that penalises segmentations for incorrect reasoning. Loss functions can also be designed to contain terms which penalise misclassifications of clinically significant regions:
\begin{equation}
\mathcal{L} = \sum_{i} \alpha_i \cdot \text{cross-entropy}(Y_i, \hat{Y}_i) + \beta \cdot \text{causal\_loss}(Y_i, \hat{Y}_i)
\end{equation}

where \(Y_i\) is the ground truth segmentation label (e.g. healthy or diseased), \(\hat{Y}_i\) is the predicted segmentation, \text{causal\_loss} penalises the model for violating causal relationships (e.g. incorrectly classifying regions based on invalid associations), and \(\alpha_i\) and \(\beta\) are weighting factors that balance the contributions of the standard cross-entropy loss and the causal loss.

\subsubsection{Instance Segmentation}


Instance segmentation relies on distinguishing individual objects in an image (e.g., lesions). CTL will improve this task by incorporating features such as lesion characteristics and their spatial relationships, thereby improving lesion detection accuracy \cite{he2017maskrcnn}.

These could be instantiations of the instance segmentation framework with CTL, trained on data-augmented samples representative of various defined by causal variables, such as variance size. In this regard, the model can use causal graphs to inform the decision process and refine relationships among detected instances.

\subsection{Predictive Modelling}

CTL provides a basis for establishing causality between imaging features and clinical outcomes in predictive modelling, allowing accurate predictions of disease outcomes. For example, CTL can be applied to predict patient responses to treatments based on pre-treatment imaging data \cite{castro2020causality, sanchez2022causal}. Predictive modelling is one of the clearest settings in which CTL must address not only covariate shift and label shift, but also concept shift. This occurs when the relationship between imaging features and clinical outcomes changes across hospitals, treatment regimes, or patient populations, for example because treatment standards, follow-up practices, or outcome definitions differ across environments. Since concept shift implies that the predictive mechanism itself is not invariant, this is one of the settings where causal transfer learning is especially necessary.

We can construct a model that directly predicts outcomes for different treatment scenarios using counterfactual reasoning. These can be formulated as follows:
\begin{equation}
\hat{Y}_{\text{treatment}} = E[Y | X, T=1] \quad \text{and} \quad \hat{Y}_{\text{control}} = E[Y | X, T=0]
\end{equation}

where \(T\) denotes treatment assignment and \( X \) represents covariates, which are the features or characteristics that influence the outcome \( Y \). In medical image analysis, \( X \) can include patient demographics, clinical data, or imaging features (e.g., tissue types, textures, or anatomical regions). Conditioning in \( X \) accounts for factors that can influence both treatment decisions and clinical outcomes, enabling more accurate counterfactual predictions and more effective intervention tailoring. By comparing these predictions, clinicians can tailor interventions more precisely based on predicted responses.

\subsection{Radiogenomics}

In radiogenomics, CTL identifies causal relationships between imaging phenotypes and genetic information to inform personalised treatment options. Genetic profile imaging uses CTL to determine how genetic variation influences disease manifestations \cite{lambin2012radiomics}. Radiogenomics most commonly faces covariate shift and conditional shift, since imaging phenotypes and genotype-linked manifestations may vary across scanners, cohorts, and acquisition settings. In addition, concept shift may arise when the biological relationship between imaging markers and genomic targets differs across populations or disease subtypes, making this an important setting for causal approaches that seek mechanistic rather than merely correlative associations.

For example, CTL can be used to identify image biomarkers for specific genetic mutations. This involves creating causal models that predict genetic outcomes based on the features of the images.
\begin{equation}
G = g(X) + \epsilon_G
\end{equation}

where \(G\) represents genetic information, \(g\) is a causal function that links imaging features \(X\) to genetic outcomes, and \(\epsilon_G\) represents the noise term or error component. The noise term captures unobserved factors or random variation that may influence the relationship between imaging features and genetic outcomes. This term is valuable for capturing the inherent uncertainty and variability in radiogenomic predictions.

\subsection{Multimodal Learning}

CTL also has an advantage in multimodal learning, in which data from different imaging modalities (MRI, CT, PET, etc.) are integrated. This supports the diagnostic process by providing a holistic view of patient conditions \cite{haq2025advancements, wang2024towards, sun2025causal, sanchez2022causal}. Multimodal learning primarily addresses covariate shift across modalities and sites, as well as conditional shift when the relationship between modality-specific observations varies across environments. In more complex clinical settings, multimodal systems may also encounter concept shift if the predictive value of one modality relative to another changes because of evolving diagnostic pathways or changing clinical decision rules.

CTL methodologies can be used to obtain joint representations of multimodal data, enabling models to learn from the causal relationships underlying different sources. One possible strategy is to use a multitask learning framework that optimises for multiple outputs but still maintains shared causal representations:

\begin{equation}
\mathcal{L}_{\text{joint}} = \sum_{i} \mathcal{L}_i + \lambda \cdot \text{causal\_regularisation}
\end{equation}

Where \(\mathcal{L}_i\) represents the individual loss for each task \(i\), such as the segmentation loss for MRI or CT data. The term \(\mathcal{L}_{\text{joint}}\) is the total loss function that combines the losses of multiple tasks. The regularisation factor \(\lambda\) balances the trade-off between the task-specific losses and the causal regularisation term. Finally, causal\_regularisation is a term that enforces consistency with causal relationships between modalities, ensuring that the model learns to maintain valid causal representations across multimodal data sources.

\subsection{Longitudinal Studies}

CTL enables the analysis of longitudinal imaging data that capture changes in patients over time. CTL continues to model causal effects of treatments over time, thus elucidating the therapeutic efficacy and disease progression \cite{nguyen2024glacial, wei2023transfer}. Longitudinal studies naturally involve label shift, as disease prevalence and outcome frequencies may change over time, and are also a key setting for concept shift, because the relationship between imaging biomarkers and outcomes can evolve during disease progression or under treatment. For this reason, longitudinal imaging is one of the most important applications in which CTL must go beyond static invariance and explicitly model changing causal mechanisms over time.

For example, one could model temporal changes in imaging biomarkers using a causal Bayesian framework and then relate these to clinical outcomes. This could be achieved with a dynamic Bayesian network, where the network updates its belief in the state of disease given new imaging data that are received over time:
\begin{equation}
P(Y_t | Y_{t-1}, X_t) = P(Y_t | \text{Parents}(Y_t)) \cdot P(X_t | Y_t)
\end{equation}

where \(Y_t\) represents the clinical state at time \(t\) and \(X_t\) represents imaging features.

\subsection{Anomaly Detection}


In anomaly detection, CTL primarily addresses covariate shift and conditional shift, since normal and abnormal image patterns may vary across scanners, populations, and acquisition environments. In rare but clinically important cases, concept shift may also arise when the definition of abnormality changes across datasets or clinical settings, requiring causal models that distinguish true pathological mechanisms from dataset-specific deviations. Thus, CTL has the potential to improve anomaly detection in medical image analysis by explicitly modelling causal relationships between image features and clinical outcomes. Instead of relying solely on statistical deviations, CTL approaches can learn the causal structure underlying normal and abnormal patterns. CTL models such as CausalGAN \cite{kocaoglu2017causalgan} and causally disentangled VAEs \cite{an2023causally} incorporate causal inference to generate and reason about images based on causal graphs. These models enable a more accurate identification of rare conditions, such as atypical lesions or rare cancers. Furthermore, structural causal modelling frameworks that integrate deep generative models with causal inference provide mechanisms to address counterfactual queries and refine latent representations based on causal relationships, as discussed in \cite{poinsot2024learning}. By explicitly modelling causal dependencies, CTL can improve the sensitivity and specificity of anomaly detection, particularly for under-represented medical conditions.

\subsection{Causal Robustness and Secure Medical Image Analysis}

Adversarial attacks and dataset shifts violate causal invariance. A model that relies on non-causal features is vulnerable not only to distribution shift but also to adversarial manipulation. CTL therefore provides a unified defence mechanism against both phenomena by enforcing that predictions depend on invariant, physiologically grounded causal mechanisms rather than spurious correlations. This application spans multiple forms of distribution shift. Adversarial perturbations and acquisition changes induce covariate shift, while changing attack patterns or dataset biases can create conditional shift. More fundamentally, concept shift may arise when the clinical meaning of image features is altered by manipulation or when the effective decision rule learned by the model ceases to reflect the true pathology-label relationship. This makes secure medical imaging one of the most challenging settings for robust causal transfer.

\subsubsection{Vulnerabilities in AI-Enabled Medical Image Analysis Systems}
The increasing integration of AI into medical image analysis has revolutionised diagnostic practice while introducing new forms of vulnerability \cite{eichelberg2020cybersecurity}. Traditional cybersecurity defences, such as network segmentation, access control, and encryption, protect the technical infrastructure of imaging systems, but leave a crucial gap in safeguarding the integrity of meaning \cite{eichelberg2020cybersecurity}. An image can be secure at the cryptographic level, yet compromised at the semantic level, where its diagnostic content no longer reflects reality. 

This semantic vulnerability arises when AI systems learn statistical patterns that do not reflect the underlying biomedical mechanisms, allowing subtle manipulations to alter the diagnostic meaning without changing the image file itself. Deep learning models that rely solely on correlations between pixels and diagnostic labels are especially susceptible to such threats, because their decisions frequently derive from non-causal statistical dependencies instead of stable causal mechanisms \cite{tjoa2020survey, rawal2025causality}. Addressing this semantic layer of security requires a shift from correlation-based pattern recognition to causality-based reasoning \cite{pearl2000causality}. 

This framework provides a means of distinguishing genuine cause–effect relationships from spurious associations \cite{pearl2000causality}. Instead of simply capturing statistical dependencies, CTL models the underlying mechanisms that generate medical images and their diagnostic outcomes \cite{pearl2000causality, scholkopf2021toward}. By representing how variables influence one another through directed relationships, CTL allows researchers to ask what would happen if a particular factor were changed and to identify which relationships are genuine and which are coincidental \cite{rawal2025causality}. In medical image analysis, this means determining whether an observed feature is directly attributable to an underlying pathological process or is a spurious association due to noise or dataset bias \cite{vlontzos2022review, mu2024explainable}.

\subsubsection{Causal Learning for Secure and Reliable Diagnostic Models}
Embedding CTL within medical image analysis fundamentally strengthens the robustness and trustworthiness of the model \cite{vlontzos2022review, jiao2024causal, rawal2025causality}. When a system models stable causal relationships in the data-generating process, deviations from these relationships may indicate noise, false associations, bias, or system instability \cite{zeng2022improving}. In medical imaging, this means that causal models may help identify predictions that conflict with expected anatomical, physiological, or pathological relationships. Therefore, such systems may be better positioned to recognise adversarial perturbations, synthetic insertions, or model-poisoning attacks that could deceive a purely statistical model \cite{tian2022confoundergan, jiao2024causal}. 

This insight parallels findings from cybersecurity research, where CTL and ensemble-based intrusion detection frameworks have been shown to improve stability under adversarial conditions by separating invariant behavioural features from spurious correlations \cite{malarkkan2025rethinking, jiao2024causal, alzubi2025detect}. Translated into the clinical domain, this principle enables medical AI to maintain consistent diagnostic reasoning even when the input is perturbed or originates from different imaging devices or institutions. 

\subsubsection{Federated and Privacy-Preserving Causal Frameworks}
Recent advances in explainable and federated medical image analysis illustrate how CTL can be operationalised as a security mechanism. Mu et al. \cite{mu2024explainable} integrated CTL into a federated learning framework enhanced with blockchain verification. In their design, causal graphs trace dependencies among local model updates, enabling the identification of anomalous or malicious contributions during aggregation. Coupled with the immutable record of the data provenance on the blockchain, this approach protects both the confidentiality and the integrity of reasoning in collaborative diagnostic models. Such architectures suggest that security in medical imaging should protect both data access and the integrity of interpretation, with causal reasoning supporting trustworthy, explainable, and semantically coherent decisions \cite{pissanetzky2016future, rawal2025causality}. 


This form of explainability further enhances the system's resilience. Conventional explainable AI techniques highlight regions in medical images that correlate with a prediction, but cannot determine whether those regions genuinely cause the outcome. CTL, on the contrary, evaluates the effect of deliberate, hypothetical changes, asking whether altering a feature would change the diagnosis \cite{pearl2000causality}. This ability to reason about interventions provides transparency that is both clinically meaningful and security-relevant, because a model’s explanation that relies on artefacts that should not affect a diagnosis can indicate possible manipulation or bias \cite{mu2024explainable}. 

At the data level, CTL also provides a novel route to privacy protection. Tian et al. \cite{tian2022confoundergan} introduced ConfounderGAN, which deliberately introduces a causal confounder to disrupt the learnable relationships between images and their diagnostic labels. The result is a set of visually authentic images that remain usable for legitimate clinical viewing but are unlearnable to unauthorised models. This causal disruption serves as a form of semantic encryption, in which sensitive information is protected not only by cryptography but also by concealing the causal pathways by which it could be inferred.

\subsubsection{Causal Anomaly Detection and System Integrity}
In this context, CTL extends cybersecurity from controlling access to ensuring the validity of reasoning. While traditional mechanisms protect where data go, CTL protects what conclusions can be drawn from them \cite{rawal2025causality}. CTL can represent established biomedical knowledge about the interactions between anatomical and physiological factors. 
When model predictions deviate from these expected causal relationships, the deviation may suggest corruption, bias, domain shift, or instability \cite{zeng2022improving}. In this sense, CTL allows model interpretability to contribute to continuous security monitoring \cite{mu2024explainable}.


This convergence of causality and security reflects the broader view that information is only trustworthy when it is causally coherent. Pissanetzky’s \cite{pissanetzky2016future} causal theory provides a useful conceptual basis for linking cybersecurity with meaning and interpretation. Although the paper takes a broader view of causality, semantics, and secure computation, its ideas can still help frame the role of interpretation and trust in medical imaging and clinical AI. From this perspective, CTL reframes medical image security as the protection not only of data, but also of the integrity of interpretation. Encryption can protect access to medical image data, while causal modelling can help safeguard the reliability, meaning, and trustworthiness of the reasoning process. 
CTL ensures that this interaction remains logically consistent and explainable, strengthening both ethical accountability and technical resilience \cite{rawal2025causality}.

CTL also informs anomaly detection and resilience in complex systems. Malarkkan et al. \cite{malarkkan2025rethinking} conceptualised causal graphs as spatio-temporal maps that can trace abnormal dependencies and isolate the sources of disruptions in cyber-physical infrastructures. Applied to medical image analysis, similar causal mapping can reveal where in the data-processing pipeline a manipulation or error originates, whether in acquisition, transmission, or inference \cite{castro2020causality}. Integrating such a CTL into federated architectures, as proposed by Mu et al. \cite{mu2024explainable}, would ensure continuous verification of data integrity in distributed clinical networks.

Together, the unifying contribution of causality lies in the link between security, privacy, and explainability under a single principle of inference \cite{rawal2025causality}. Encryption and blockchain secure the transport and storage of information \cite{mu2024explainable}, but they do not guarantee that model outputs are clinically meaningful, reliable, or intelligible. CTL addresses this gap by encouraging AI systems to base their conclusions on mechanisms consistent with medical reality. By embedding CTL into medical image analysis, researchers can build systems that are not only accurate but also more robust, interpretable, and secure by design \cite{mu2024explainable, tian2022confoundergan}.


\subsection{Imaging Protocol Optimisation with Causal Inference}

Finally, CTL can optimise medical image analysis workflows by identifying factors that affect image quality and diagnostic accuracy. Analysing the causal relationships between imaging parameters, patient demographics, and diagnostic outcomes informs imaging protocols that are clinically meaningful and empirically grounded \cite{baniasadi2022dbsegment, subbaswamy2018counterfactual, monsell2023statistical, gnanakalavathy2025capri, zech2018variable, gichoya2022ai, zhuang2021multipleshooting, vlontzos2022causal}.
Thus, imaging protocol optimisation primarily addresses covariate shift, since protocol changes directly alter the observed image distribution, and conditional shift, when protocol-dependent image characteristics modify the apparent manifestation of pathology. In some cases, protocol changes may also contribute to concept shift if they alter the effective relationship between measured image features and downstream diagnostic decisions, reinforcing the need for causal analysis when comparing protocols across institutions.

For example, in breast cancer screening, CTL might help identify the most effective imaging protocols based on specific patient demographics (e.g., age, breast density) and prior medical history, thereby improving diagnostic outcomes. This can be achieved by developing causal models that predict diagnostic outcomes across different imaging strategies, thereby enabling workflows to be adapted to optimise patient care. Additionally, CTL could support the standardisation of diagnostic protocols across institutions, ensuring consistent diagnostic accuracy while reducing disparities in care  \cite{chen2021algorithm, mueller2021causes}.

\section{Causal Transfer Learning Applied in Medical Image Analysis}    
    {In the context of medical image analysis, CTL offers several key advantages, which become particularly evident when examining its practical applications in clinical settings. That is, CTL improves generalisation across domains, allowing models to adapt more effectively to new imaging devices, centres, or modalities. Additionally, CTL enhances robustness to domain shift and heterogeneity, addressing the performance degradation often caused by differences in imaging protocols, populations, and devices. It leverages causal inference techniques, such as back-door adjustment and style transfer, to mitigate these effects \cite{wu2024causal}. Furthermore, given the high cost of acquiring annotated medical image analysis datasets, CTL facilitates efficient learning from limited annotated data. By incorporating causal structure, CTL supports few-shot and self-supervised learning, thereby reducing annotation burden \cite{ouyang2022self}. Another significant benefit is its capacity to enhance clinical trustworthiness. By explicitly defining the causal relationships between imaging features (such as tissue textures or anatomical structures) and pathological outcomes (e.g., tumour growth or disease progression), CTL enhances model interpretability, reduces biases, and promotes more reliable deployment in clinical settings \cite{jones2023no, jones2024causal}.

    The eleven case studies presented in Sections 6.1–6.11 were selected from the included papers to provide representative coverage across the taxonomy dimensions established in this review: they span four imaging modalities (fundus, MRI, CT, chest X-ray), four task types (segmentation, classification, reconstruction/enhancement, semi-supervised learning), and four shift types (covariate, conditional, scanner-induced, and protocol-induced shift). Selection prioritised studies that (i) explicitly state their causal framework (e.g., SCM, counterfactual generation, IRM), (ii) report results on publicly available benchmark datasets enabling comparison, and (iii) have been published in peer-reviewed venues indexed by the databases searched.

    Recent studies have demonstrated the broader application of causal learning in medical image analysis. For example, MACAW \cite{vigneshwaran2024macaw} introduces a causal generative model for medical image generation. In contrast, Semi-Supervised Learning for Deep Causal Generative Models \cite{ibrahim2024semi} discusses causal reasoning in data-efficient representation learning. The CausCLIP \cite{li2024causclip} applies causal reasoning to visual-language models for few-shot echo-cardiographic reporting quality assessment. These training examples highlight the broader value of causal learning in medical image analysis and support the development of robust models. Additionally, \cite{zhu2025counterfactual} developed a structured causal model to generate clinically meaningful counterfactual images for lung disease diagnosis, illustrating the potential of causal models in creating alternative clinical scenarios.
    
    In this section, we review state-of-the-art (SOTA) CTL methods applied to medical image analysis, categorised by task type (segmentation, classification, reconstruction, domain adaptation/generalisation), and summarise the associated datasets, causal frameworks (e.g., structural causal models, interventions, invariance), results, and clinical relevance. We then explore key challenges and solutions, highlighting applications across various imaging modalities (e.g., fundus imaging, MRI, CT, histopathology), ultimately illustrating how CTL is transforming the development of machine learning models for clinical imaging and outlining future research directions. Additionally, Table \ref{challenges_solutions} reviews nine challenges and their corresponding solutions in causal transfer learning.}

\begin{table}[H]
\centering
\scriptsize
\caption{Key challenges in medical image (MI) analysis and causality-inspired solutions}
\label{challenges_solutions}

\begin{tabular}{@{}>{\raggedright\arraybackslash}p{4cm}
                >{\raggedright\arraybackslash}p{11cm}@{}}
\toprule
\textbf{Challenge} & \textbf{Causality-Inspired Solution(s)} \\
\midrule
\multicolumn{2}{l}{\textbf{(A) Data limitations}} \\
\midrule
Limited training data & One-shot / few-shot causal learning \cite{carloni2023causality} \\
Lack of labeled target-domain data & Self-supervised learning \cite{li2025causal, qiu2023causality} \\
\midrule
\multicolumn{2}{l}{\textbf{(B) Domain shift and heterogeneity}} \\
\midrule
Style discrepancies between imaging domains & Fourier-based style transfer \cite{li2025causal} \\
Differences in feature space between domains & Cross-domain contrastive learning with adversarial training \cite{li2025causal}; contrastive feature alignment with source prototypes \cite{qiu2023causality} \\
Cross-modality domain shift & Diffusion models for modality-invariant features \cite{chen2025generalizable} \\
Source-free domain adaptation & Prototype-based contrastive feature alignment and causal interventions \cite{qiu2023causality} \\
Variations in intensity and texture & Causality-driven augmentations, including GIN and IPA \cite{ouyang2022causality, chen2025generalizable, qiu2023causality} \\
\midrule
\multicolumn{2}{l}{\textbf{(C) Bias and spurious effects}} \\
\midrule
Confounding factors in feature extraction & Backdoor adjustment \cite{li2025causal, qiu2023causality} \\
Spurious correlations in object appearance & Data-augmentation-based causal intervention \cite{ouyang2022causality, qiu2023causality} \\
\bottomrule
\end{tabular}

\end{table}

    \subsection{Causal Inference-Based Self-Supervised Cross-Domain Fundus Image Segmentation}
    
        \begin{figure}[htbp]
        \centering
            \includegraphics[width=0.5\textwidth]{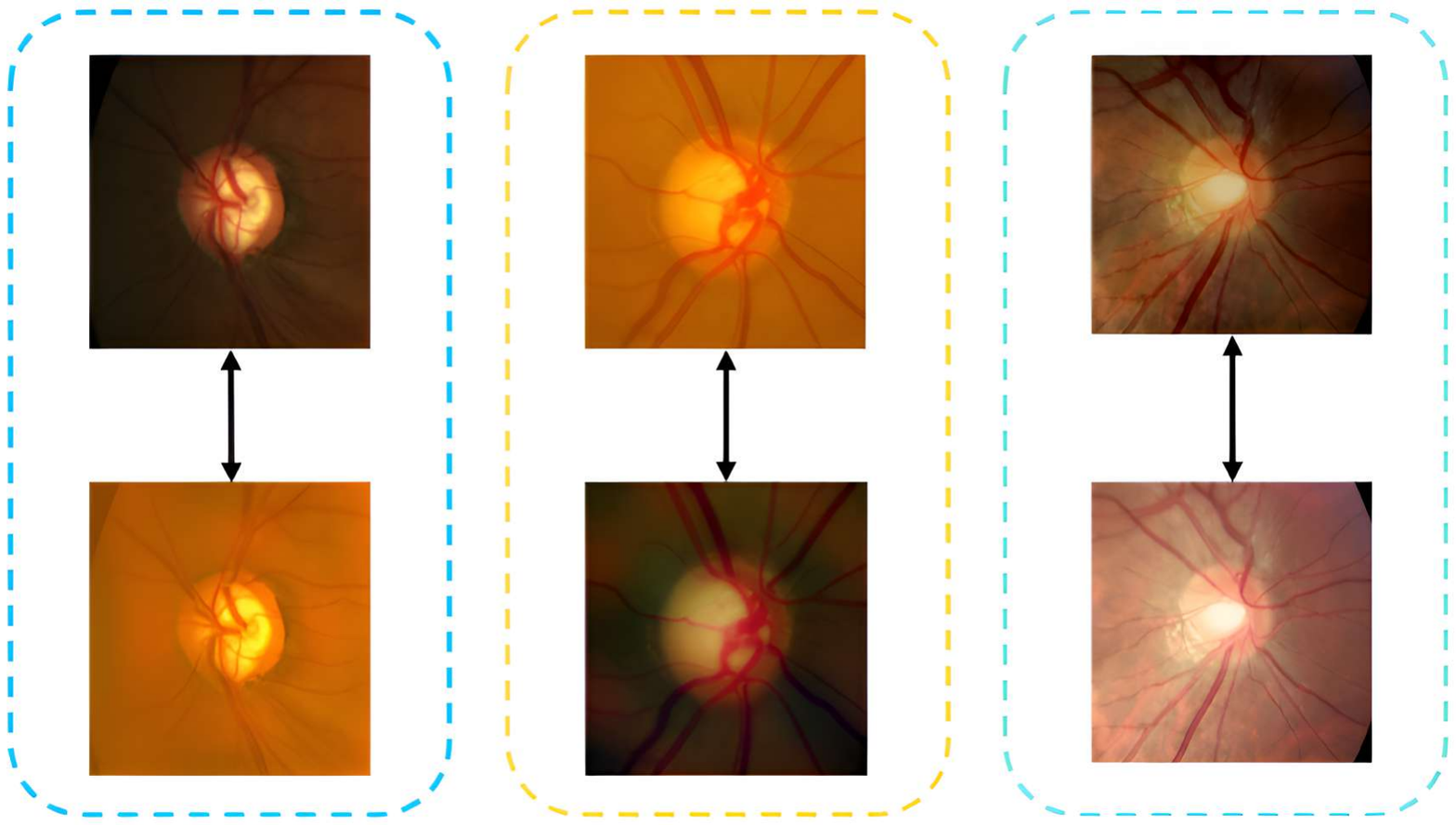}
            \caption{Illustration of the domain-shift problem of Fundus in three samples \cite{li2025causal}.}
            \label{fig:style_shift}
        \end{figure}

        This case study addresses \textbf{covariate shift} induced by differences in fundus camera hardware across clinical sites, representing the most common form of domain shift in ophthalmic imaging.

        {Glaucoma is one of the leading causes of irreversible blindness, with early detection being vital for preventing vision loss. Cup-to-disc ratio (CDR) estimation from fundus images is a key diagnostic criterion for glaucoma. However, accurate CDR estimation is complicated by variations across imaging devices. Different fundus cameras (e.g., Zeiss Bisucam vs Canon CR-2) produce images with distinct styles (e.g., contrast, resolution, and colour balance), leading to domain shifts between the training and target data. This domain shift significantly affects the performance of machine learning models, which typically rely on labelled datasets for training. When target-domain labels are unavailable, this challenge becomes even more pronounced. The challenge of domain shift in medical image analysis is particularly evident in the segmentation of fundus images for glaucoma diagnosis, as shown in Figure \ref{fig:style_shift}, where performance degrades due to style discrepancies between the source and target domains. This section presents a method developed by \cite{li2025causal} that uses Causal Self-Supervised Networks (CSSN) to tackle this issue and achieve robust cross-domain segmentation.}

        \subsubsection{Experimental Dataset}
            {The study utilises the REFUGE dataset \cite{orlando2020refuge} as the source domain, which includes 400 annotated training images from a Zeiss Bisucam 500, along with 400 images from the Canon CR-2 for validation. The target domains comprise three datasets with varying imaging characteristics: REFUGE Validation/Test \cite{orlando2020refuge}, Drishti-GS \cite{sivaswamy2015comprehensive}, and RIM-ONE-r3-all \cite{fumero2011rim}, which simulate real-world imaging variations. These datasets represent diverse clinical settings and camera types, thereby simulating real-world scenarios in which models must perform well across a wide range of imaging devices. This dataset setup emphasises the domain shift problem in medical image segmentation, as the same eye may appear drastically different across devices.}
            
        \subsubsection{Methodology}       
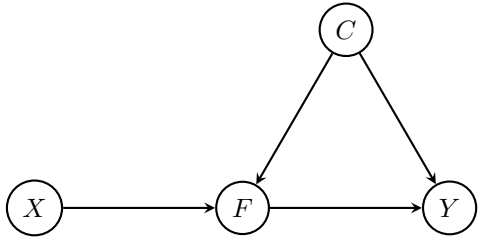
\begin{figure}[htbp]
    \centering
    \begin{tikzpicture}[->,>=stealth, node distance=2cm, thick]
        \node[circle,draw] (X) {$X$};
        \node[circle,draw,right=of X] (F) {$F$};
        \node[circle,draw,right=of F] (Y) {$Y$};
        \node[circle,draw,above=2cm of $(F)!0.5!(Y)$] (C) {$C$};

        \draw (X) -- (F);
        \draw (C) -- (F);
        \draw (C) -- (Y);
        \draw (F) -- (Y);
    \end{tikzpicture}
    \caption{SCM for mitigating domain style interference \cite{li2025causal}.}
    \label{fig:scm}
\end{figure}

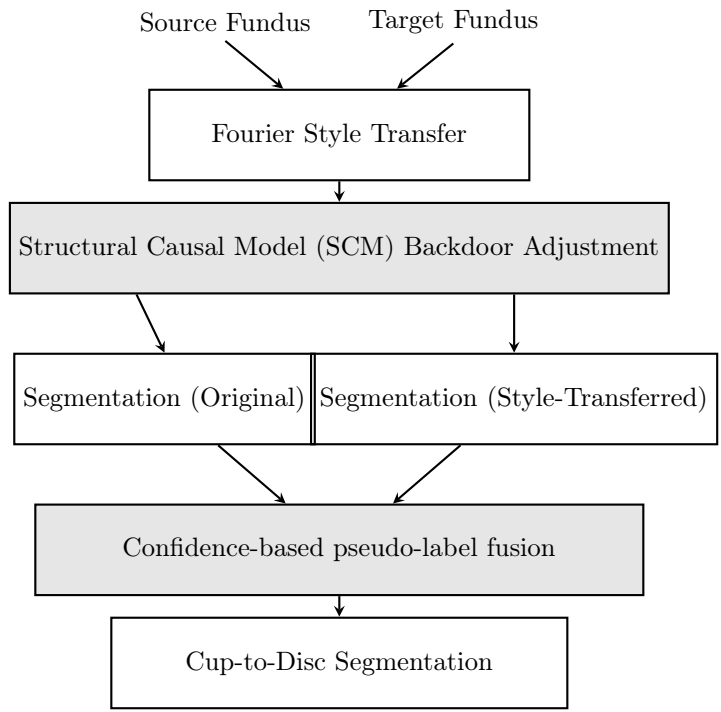
\begin{figure}[htbp]
        \centering
        \begin{tikzpicture}[>=stealth, thick]

            \node[] (source) {Source Fundus};
            \node[] (target) at ($(source) + (3cm,0)$) {Target Fundus};

            \node[draw, minimum width=5cm, minimum height=1.2cm](fourier) at ($(source)!0.5!(target) + (0,-1.5cm)$) {Fourier Style Transfer};

            \node[draw, fill=gray!20, rectangle, minimum width=5cm, minimum height=1.2cm](scm) at ($(fourier) + (0,-1.5cm)$) {Structural Causal Model (SCM) Backdoor Adjustment};

            \node[draw, rectangle, minimum width=2.5cm, minimum height=1.2cm] (segA) at ($(scm) + (-2.3cm,-2cm)$) {Segmentation (Original)};
            \node[draw, rectangle, minimum width=2.5cm, minimum height=1.2cm] (segB) at ($(scm) + (2.3cm,-2cm)$) {Segmentation (Style-Transferred)};

            \node[draw, fill=gray!20, minimum width=8cm, minimum height=1.2cm] (fusion) at ($(scm) + (0,-4cm)$) {Confidence-based pseudo-label fusion};

            \node[draw, rectangle, minimum width=6cm, minimum height=1.2cm] (output) at ($(fusion) + (0,-1.5cm)$) {Cup-to-Disc Segmentation};

            \draw[->] (source.south) -- (fourier);
            \draw[->] (target.south) -- (fourier);
            \draw[->] (fourier) -- (scm);
            \draw[->] ([xshift=16.8mm] scm.south west) -- (segA.north);
            \draw[->] ([xshift=23mm]scm.south) -- (segB);
            \draw[->] (segA) -- (fusion);
            \draw[->] (segB) -- (fusion);
            \draw[->] (fusion) -- (output);

            \end{tikzpicture}
            \caption{Simplified Causal Self-Supervised Network (CSSN) \cite{li2025causal}.}
            \label{fig:cssn}
            \end{figure}

            {The CSSN framework begins by constructing a Structural Causal Model (SCM) to address the domain shift problem in medical image segmentation. In this framework, the target-domain images ($X$) are first extracted into feature maps ($F$), from which pseudo-labels ($Y$) are derived. However, the process is complicated by the presence of a domain style confounder ($C$), a factor that introduces spurious style effects that influence both feature extraction and pseudo label generation, resulting in bias in $Y$ through backdoor paths (as illustrated in Figure ~\ref{fig:scm}). To mitigate these biases, the framework employs a Fourier-based approach to swap low-frequency components between source and target images, thereby creating style-augmented variants. A dual-path segmentation network processes both the original and style-transferred inputs. These predictions are then merged using a confidence-based pseudo-label fusion strategy, providing reliable supervision for segmentation tasks. To further improve cross-domain generalisation, the method integrates adversarial training for feature alignment and employs cross-domain contrastive learning to preserve structural consistency, ensuring robust performance across diverse domains \cite{li2025causal, pawlowski2020deep, glymour2019review}.} Figure \ref {fig:cssn} illustrates the simplified architecture of the Causal Self-Supervised Network (CSSN).

        \subsubsection{Result}
            {The CSSN model demonstrated impressive performance across the three target datasets: On Drishti-GS, the model achieved DI$_{\text{cup}}$ = 0.876 and DI$_{\text{disc}}$ = 0.971, with a $\delta$ of 0.081; On RIM-ONE-r3, the model scored DI$_{\text{cup}}$ = 0.818 and DI$_{\text{disc}}$ = 0.922, with a $\delta$ of 0.083; and On REFUGE, the model achieved DI$_{\text{cup}}$ = 0.885 and DI$_{\text{disc}}$ = 0.958, with a $\delta$ of 0.049 \cite{li2025causal}. These results demonstrate the model’s ability to generalise across different imaging devices and clinical settings.}
        
        \subsubsection{Clinical Relevance and Future Directions}
            {The CSSN framework provides an effective solution for cross-domain segmentation in fundus imaging. Its key advantages lie in its robustness to domain shifts and its ability to perform well on unlabelled target domains using self-supervised learning. This is especially valuable in clinical settings where annotated data are scarce. In addition, the Fourier-based style transfer and confidence-based pseudo-label fusion strategies help make the model data-efficient and scalable across multiple imaging devices and clinical contexts. The proposed approach significantly enhances early glaucoma detection by reducing domain-specific biases introduced by different fundus cameras. Looking ahead, future work could extend this approach to other multimodal segmentation tasks (e.g., MRI and CT scans) and integrate it into clinical decision support systems to support more comprehensive diagnostic workflows.}

    \subsection{Generalisable Single-Source Cross-Modality Segmentation via Invariant Causality}
        

        This case study addresses \textbf{conditional shift} arising from systematic differences in image appearance across modalities (CT vs \ MRI), where the same anatomical structure manifests differently depending on the imaging device and acquisition sequence. This form of shift is especially severe in single-source generalisation settings, where no labelled target-domain data are available. Effective segmentation across multiple modalities (e.g., CT, MRI, and PET) is consequently challenging due to the resulting differences in contrast, resolution, and structural representation. This challenge is especially pronounced when attempting to generalise from a single-source modality (e.g., CT) to unseen target modalities (e.g., MRI) with limited or no labelled data.  Traditional machine learning models often struggle with this type of generalisation, as they tend to rely heavily on modality-specific features, which can be influenced by imaging styles rather than domain-invariant anatomical structures. To overcome these limitations, the study by \cite{chen2025generalizable} employs an SCM, which distinguishes between anatomical content (the true structural features of the image, such as tissue types and organ shapes) and modality-specific style (the appearance of the image, influenced by factors like contrast, resolution, and scanner settings). This causal framework uses controlled diffusion interventions to modify imaging styles while preserving the structural consistency of anatomical features across modalities, thus enabling the model to generalise effectively from a single-source modality to new, previously unseen imaging types. In the next section, we’ll examine the model’s architecture in greater detail, explaining how these causal interventions ensure reliable segmentation performance across diverse imaging modalities.

        \subsubsection{Methodology}
            {The data creation process is modelled using a Structural Causal Model (SCM), which separates the image into anatomical content and modality-specific style. Controlled diffusion models are used to adapt the style, ensuring consistency across modalities \cite{fang2024data}. The principle of intervention-augmentation equivariance is applied to maintain consistent network predictions across varying styles. The segmentation network is then trained on both original and style-modified images, using loss functions to preserve anatomical structure and ensure generalisation \cite{chen2025generalizable}.}

        \subsubsection{Result}
            {The model was evaluated on three cross-modality segmentation tasks: Abdominal Segmentation (CT → MRI), where it achieved Dice scores of 86.20\%, demonstrating strong performance in adapting to a new modality; Lumbar Spine Segmentation (MRI → CT), where it surpassed prior methods with an improvement of 3.13\% in Dice scores; and Lung Segmentation, which achieved a 78.79\% average Dice score, outperforming baseline methods by more than 10\%. These results underscore the model's ability to generalise effectively from a single-source modality (e.g., CT) to unseen target modalities (e.g., MRI), outpacing traditional methods on tasks where cross-modality generalisation is crucial. 
            }
            
        \subsubsection{Clinical Relevance and Future Directions}
            {The proposed causality-inspired model introduces a novel approach to cross-modality segmentation by focusing on domain-invariant features. This work significantly advances traditional methods that often overfit to modality-specific features and fail to generalise across different imaging devices. By incorporating causal reasoning to separate anatomical content from modality-specific style, the model achieves robust performance even in the absence of data from the target modality. Its key contributions include strong performance in cross-modality image segmentation (e.g., CT-to-MRI), which is essential for medical image analysis systems that must operate across diverse imaging modalities. The integration of causal interventions ensures that the model focuses on features that accurately represent the anatomy, thereby minimising the impact of modality-specific biases. Looking ahead, this approach could be extended to multimodal segmentation tasks (e.g., integrating CT, MRI, and PET for comprehensive diagnostic systems) or adapted for source-free domain adaptation, enabling the model to generalise across multiple domains without access to source-domain data. Additionally, scaling the approach to handle larger datasets and enabling real-time clinical applications could further enhance its utility across diverse medical settings.}

            Building on style-invariant representations, the following section examines segmentation tasks and enhances generalisation through targeted causal data augmentation rather than via diffusion-based style modifications. The subsequent work builds on these causal principles to advance semi-supervised learning for medical image segmentation, particularly in scenarios with limited labelled data.

    \subsection{Causality-Inspired Single-Source Domain Generalisation for Segmentation}


        This case study addresses \textbf{covariate shift} due to acquisition-specific intensity and texture variations across scanners, protocols, and patient populations, using causal data augmentation to promote domain-invariant representations. Single-source domain generalisation is a significant challenge in medical image analysis, where models often overfit to modality-specific characteristics, such as intensity, texture, and background features, that do not generalise across diverse clinical environments. {These spurious correlations degrade performance when the model encounters domain shifts due to variations in scanners, acquisition protocols, or patient populations. To mitigate this, \cite{ouyang2022causality} proposes a causality-inspired framework that incorporates causal data augmentation and explicitly decouples anatomical structures (causal factors) from modality-specific appearance variations (non-causal nuisance factors). By generating style-diverse synthetic samples that preserve anatomical integrity, this method encourages models to focus on domain-invariant structural features, enhancing segmentation robustness and generalisation across unseen domains while minimising bias introduced by acquisition-specific inconsistencies.}

        \subsubsection{Methodology}
\begin{figure*}[t]
    \centering
    \resizebox{\textwidth}{!}{%
    \begin{tikzpicture}[>=latex, node distance=2mm]

        \node (input) {\includegraphics[width=2cm]{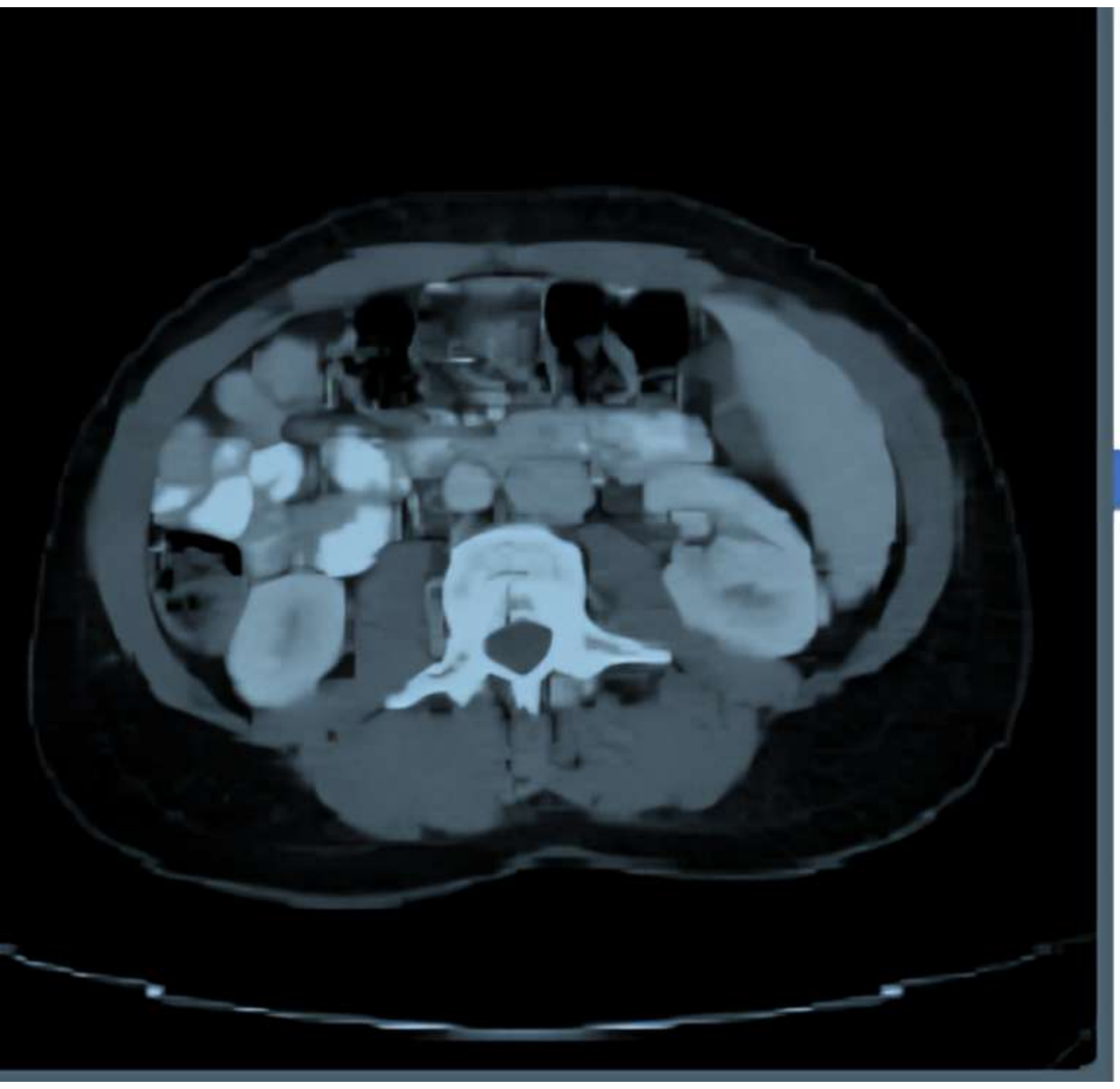}};
        \node[below=1mm of input] {Input Image};

        \node[draw, rectangle, minimum width=2.6cm, minimum height=1.2cm, right=3mm of input] (cnn)
        {\shortstack{Shallow CNN \\ (Random Weights + LeakyReLU)}};
        \node[below=1mm of cnn]
        {\shortstack{Global Intensity \\ Non-Linear Augmentation}};

        \node[draw, circle, minimum size=7mm, right=6mm of cnn] (plus) {+};

        \node[draw, rectangle, fill=green!20, minimum width=1.8cm, minimum height=1.2cm, right=4mm of plus] (norm)
        {\shortstack{Re-normalisation}};

        \node[right=4mm of norm] (output) {\includegraphics[width=1.4cm]{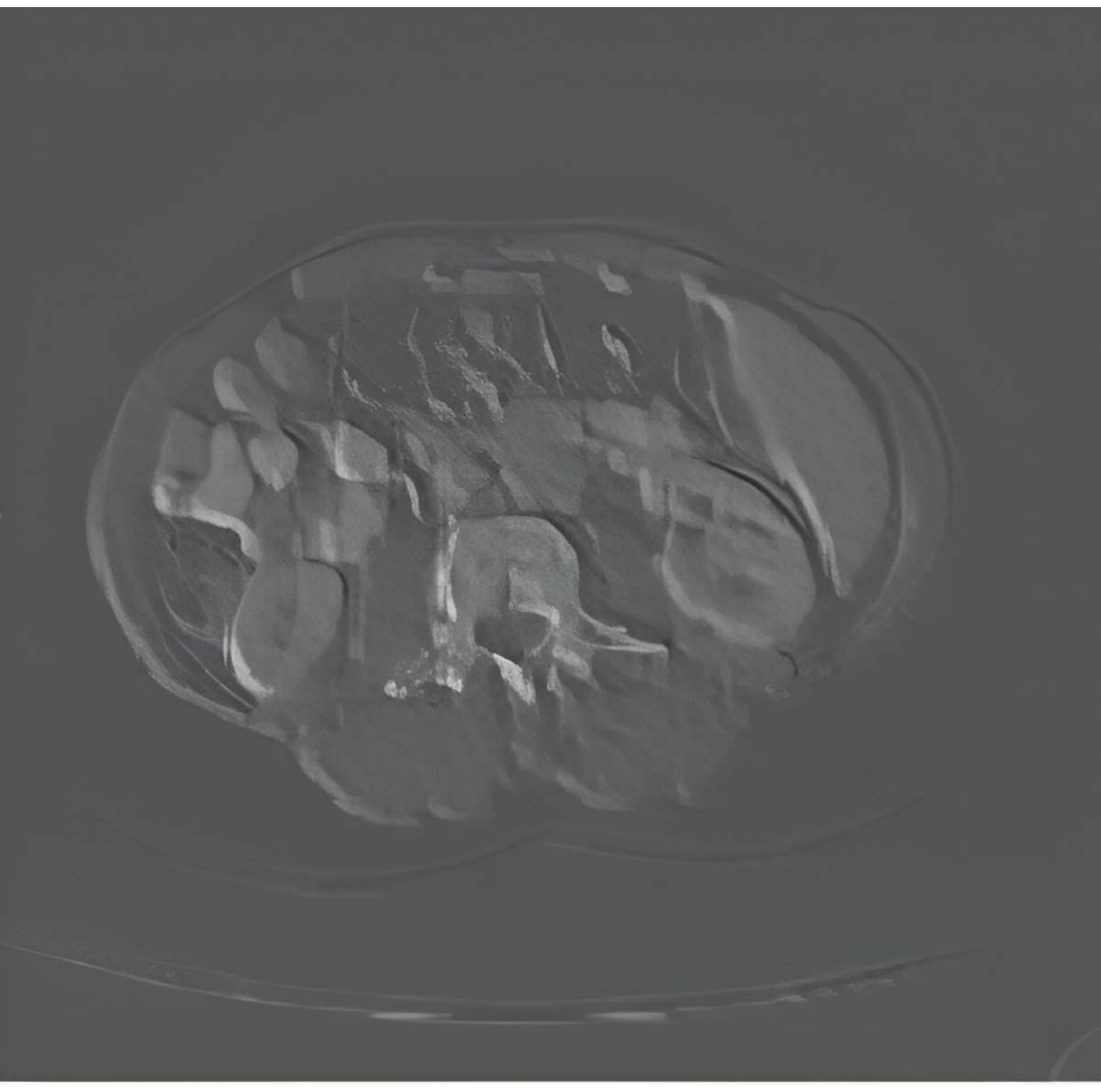}};
        \node[below=1mm of output] {GIN Augmented Image};

        \draw[->, thick] (input) -- (cnn);
        \draw[->, thick] (cnn) -- (plus);
        \draw[->, thick] (plus) -- (norm);
        \draw[->, thick] (norm) -- (output);

        \draw[->, thick, dashed]
            (input.east) .. controls +(1,1.5) and +(-1.5,1.5) .. (plus.north west);

    \end{tikzpicture}
    }
    \caption{Global Intensity Non-linear Augmentation (GIN) \cite{ouyang2022causality}.}
    \label{fig:GIN}
\end{figure*}

            {The proposed method comprises two main components: (1) Global Intensity Nonlinear Augmentation (GIN): GIN, as shown in Figure \ref{fig:GIN}, introduces intensity-based perturbations using shallow, randomly weighted CNNs combined with nonlinear activations (e.g., LeakyReLU). These perturbations simulate realistic variations in imaging appearance, such as altered contrast, scanner noise, or illumination differences, without distorting anatomical structures. After perturbation, a renormalisation step ensures intensity distributions remain physiologically plausible. The goal is to expand training data by generating diverse, anatomically consistent images that simulate the effects of different acquisition devices, patient populations, and imaging parameters. Importantly, GIN perturbations follow the causal principle that appearance is a non-causal variable, and modifying it should not alter the segmentation-relevant anatomy. (2) Interventional Pseudo-Correlation Augmentation (IPA): IPA targets confounding correlations between background textures and organ appearance by independently resampling foreground and background features. This intervention simulates the breaking of spurious dependencies introduced during image acquisition. By decorrelating these factors, the segmentation model is forced to rely on anatomical shapes and boundaries, the true causal signals, rather than acquisition artefacts. Both GIN and IPA are modular and can be applied on top of any existing segmentation architecture. During training, the network simultaneously processes original and augmented images, encouraging invariance and robustness to style, noise, and acquisition variability.}

        \subsubsection{Result}
        {The proposed causality-driven augmentation framework was evaluated across three cross-domain segmentation benchmarks: the abdominal MRI/CT dataset (Dice score: 86.3), cardiac MRI (multi-centre) dataset (Dice score: 85.0), and prostate multi-domain dataset (Dice score: 70.4) \cite{ouyang2022causality}. These results surpass traditional augmentation strategies and outperform existing domain-generalisation methods, particularly in scenarios where the target domain differs significantly from the source in intensity, acquisition parameters, or scanner type.}
        
        \subsubsection{Clinical Relevance and Future Directions}
        {This work represents a significant advancement in single-source domain generalisation for medical image segmentation by incorporating causal principles into data augmentation. The authors directly address the root cause of poor generalisation: spurious dependencies arising from acquisition-specific features. Unlike conventional augmentations, GIN and IPA generate physiologically valid variations that meaningfully stress-test the model’s invariance to style, contrast, and noise. The framework's strength lies in its simplicity, modularity, and scalability, making it easily incorporable into any segmentation pipeline without requiring target-domain data. This is especially valuable in clinical workflows, where data sharing restrictions and privacy concerns limit access to multi-domain datasets. Future research could integrate this causal augmentation strategy with diffusion models, generative priors, or source-free domain adaptation, potentially enabling more robust generalisation in large, heterogeneous clinical systems. Additionally, extending the framework to multi-organ, multi-modality, or 3D volumetric segmentation tasks presents promising directions for expanding clinical impact.}

    \subsection{CauSSL: Causality-inspired Semi-supervised Learning for Medical Image Segmentation}
        
        This case study addresses \textbf{covariate shift} arising from differences in intensity and acquisition protocols across MRI and CT datasets, compounded by the practical constraint of severely limited labelled data. When labelled examples are scarce, models trained under covariate shift cannot rely on large supervised signals to compensate for distributional mismatch, making causal semi-supervised learning the natural methodological response.
        {Building on this motivation, \cite{miao2023caussl} presents CauSSL, a novel causality-inspired semi-supervised learning (SSL) approach aimed at enhancing medical image segmentation. Despite the empirical success of semi-supervised learning in medical image segmentation tasks, a major challenge remains: the theoretical understanding of its effectiveness, particularly the mechanisms by which unlabeled data improve performance.  Here, the authors propose a causal diagram to provide a theoretical foundation for SSL methods and to address concerns about algorithmic independence between networks in co-training frameworks. By focusing on algorithmic independence and employing a min-max optimisation approach, CauSSL improves the performance of existing SSL methods, particularly in medical image segmentation tasks with limited labelled data. The study demonstrates the effectiveness of CauSSL across 2D and 3D network architectures, providing significant improvements over SOTA methods on three public medical image segmentation datasets.}
        
        \subsubsection{Experimental Dataset}
            {The CauSSL framework was evaluated on three widely-used medical image segmentation datasets: the Automatic Cardiac Diagnosis Challenge (ACDC) dataset \cite{bernard2018deep}, which focuses on the segmentation of cardiac structures from MRI data; the Pancreas-CT dataset \cite{clark2013cancer, roth2016data, roth2015deeporgan}, which involves pancreas segmentation from CT images; and the Multimodal Brain Tumor Segmentation Challenge 2019 (BraTS'19) dataset \cite{bakas2017advancing, bakas2018identifying, baid2021rsna, menze2014multimodal}, 
            which contains multi-modal MRI data for brain tumour segmentation. These datasets cover diverse image modalities and segmentation tasks, enabling assessment of CauSSL's generalisability and performance across different types of medical image analysis data.
            }
            
        \subsubsection{Methodology}
            {The core innovation of CauSSL lies in its causality-inspired framework for SSL, which involves a causal diagram that introduces intermediate variables, such as pseudo-labels or predictions from another network, to better explain the success of SSL methods in segmentation. The authors argue that algorithmic independence between networks, in which each network's predictions do not unduly influence the other’s learning process, is crucial for improving performance. To achieve this, CauSSL employs a min-max optimisation strategy to maximise the independence between the networks, thereby enhancing overall segmentation performance. The framework quantifies network independence using the Minimum Description Length (MDL) principle, specifically applied to deep convolutional networks. The networks are optimised by minimising the loss on both labelled and unlabelled data while maximising the algorithmic independence between the two networks. This approach is integrated into popular SSL frameworks, such as co-training, and the authors demonstrate its effectiveness by improving both segmentation quality and efficiency.}
            
        \subsubsection{Result}
            {The proposed CauSSL framework was tested against SOTA SSL methods on three different datasets and two distinct network architectures (2D U-Net and 3D V-Net). The results highlight significant performance gains, with CauSSL consistently outperforming standard SSL methods such as Mean Teacher (MT) and co-training (CPS, MC-Net+), particularly when labelled data were limited. On the ACDC dataset, CauSSL improved the Dice Similarity Coefficient (DSC) by 1.01\% with 10\% labelled data and by 0.74\% with 20\% labelled data. On the Pancreas-CT dataset, it achieved a 4.71\% improvement in DSC over MC-Net+ with only six annotated volumes. On the BraTS'19 dataset, CauSSL showed a 1\% improvement in DSC over CPS and MC-Net+. By incorporating a network-independence constraint, CauSSL significantly reduced algorithmic dependence, resulting in improved segmentation performance on the ACDC and Pancreas-CT datasets, where the min-max optimisation strategy further enhanced network independence. Furthermore, the quality of the segment, particularly for challenges such as tumour segmentation on BraTS'19 and pancreas segmentation on Pancreas-CT, was notably higher, as evidenced by improved Jaccard Index (JC) and Hausdorff Distance (95HD) metrics.}
            
        \subsubsection{Clinical Relevance and Future Directions}
            The CauSSL framework has significant clinical implications, particularly for medical image segmentation tasks where labelled data are scarce and costly to obtain. By improving data efficiency, CauSSL enables better segmentation performance, limited labelled data, making it highly relevant for clinical segmentation. Professional annotation is both time-consuming and expensive. Future directions for CauSSL include extending it to other medical image modalities, such as CT scans and MRI, for tasks including organ segmentation and lesion detection. Additionally, integrating CauSSL with other deep learning methods, such as multi-task learning (MTL) or domain adaptation, could enhance model generalisation across various hospital centres. Further refinement of the statistical quantification of network independence would make the framework more adaptable to a broader range of architectures and data types. Finally, testing CauSSL in real-world clinical environments with diverse data sources could validate its effectiveness and robustness in dynamic healthcare settings.

\subsection{Multimodal Causality-Driven Representation Learning for Generalizable Medical Image Segmentation}

This case study primarily addresses \textbf{covariate shift}, where the input distribution changes across clinical environments because of differences in equipment, procedure artefacts, and imaging modes. At the same time, the underlying lesion semantics remain stable. The method is designed for multi-source-to-single-source domain generalisation across clinical sites, with causal interventions used to suppress domain-specific confounders and recover domain-invariant lesion representations.

Medical image segmentation often fails to generalise across clinical sites because lesion appearance is confounded by acquisition-specific factors such as endoscopic hardware, illumination conditions, and procedural artefacts. These confounders alter the observed image distribution without necessarily changing the underlying pathology, making this setting a canonical example of covariate shift in medical imaging. To address this problem, \cite{liang2026multimodal} propose \emph{Multimodal Causality-Driven Representation Learning} (MCDRL), a framework that integrates causal inference with vision-language modelling for domain-generalised medical image segmentation. The key idea is to explicitly model and remove domain-specific confounders while preserving anatomically relevant lesion structure, thereby improving generalisation to unseen sites. The authors motivate this design by noting that existing domain-generalisation approaches often learn domain-invariant representations only implicitly, whereas MCDRL uses causal intervention to target the confounders themselves \cite{liang2026multimodal}.

\subsubsection{Experimental Dataset}
The study evaluates MCDRL on five clinical sites spanning two endoscopic application domains. The laryngoscopy evaluation uses the \emph{Laryngoscope8} dataset (Site B), comprising 3,533 annotated images with segmentation masks for vocal tract pathologies \cite{yin2021laryngoscope8}. The laparoscopy/colonoscopy evaluation further uses \emph{CVC-ClinicDB/CVC-612} (Site C, 612 images) \cite{bernal2015wm}, \emph{ETIS} (Site D, 196 images) \cite{silva2014toward}, and \emph{Kvasir-SEG} (Site E, 1,000 images) \cite{jha2019kvasir}, all collected from different medical centres. The experimental setup is a multi-source-to-single-source domain-generalisation protocol, in which the model is trained on multiple sites and tested on an unseen target site, thereby directly evaluating cross-site generalisability under domain shift \cite{liang2026multimodal}.

\subsubsection{Methodology}
MCDRL combines multimodal representation learning with explicit causal intervention. First, the framework leverages CLIP-style visual-language alignment to identify candidate lesion regions and construct a \emph{confounder dictionary} via text prompts representing domain-specific variations. Second, it trains a \emph{causal intervention network} that uses this dictionary to identify and suppress the influence of confounders while preserving anatomical structure relevant to segmentation. In the authors’ formulation, domain shift arises from confounding factors such as equipment differences, procedure artefacts, and imaging modes, and the model aims to intervene on these nuisance factors rather than align feature spaces statistically. This makes the method particularly relevant to CTL, because it combines multimodal representation learning, domain generalisation, and causal intervention in a unified segmentation pipeline \cite{liang2026multimodal}.

\subsubsection{Result}
MCDRL consistently outperforms competing methods across sites, lesion types, and backbone architectures. Using a ResNet-50 backbone, the method achieves an average Dice of 78.6 compared with 76.7 for BiomedCoOp and 72.1 for the baseline. The authors further report that, with a ViT-L/14 backbone, MCDRL reaches an average mDice of 81.6, corresponding to a 6.5-point improvement over the baseline and a 2.0-point improvement over the strongest competing method. Across lesion categories, the method improves average mDice from 78.5 to 82.4 relative to BiomedCoOp, with especially large gains for nodules, where the reported improvement is 11.4 points over the baseline. These results indicate that causal suppression of confounders improves segmentation robustness across unseen clinical sites \cite{liang2026multimodal}.

\subsubsection{Clinical Relevance and Future Directions}
MCDRL is clinically relevant because it addresses one of the most persistent barriers to real-world deployment: the degradation of segmentation performance when endoscopic data are acquired under different clinical conditions. By explicitly targeting confounders rather than relying only on implicit invariance, the method offers a more principled route to robust cross-site segmentation. Its integration of visual and textual information also suggests that CTL is increasingly moving toward multimodal and foundation-model-based pipelines. Looking ahead, this approach could be extended to other multi-site segmentation settings, including MRI and CT, and to broader multimodal workflows in which textual reports, acquisition metadata, and image content jointly inform causal domain generalisation \cite{liang2026multimodal}.
\subsection{MuCALD-SplitFed: Causal-Latent Diffusion for Privacy-Preserving Multi-Task Split-Federated Medical Image Segmentation}

This case study primarily addresses \textbf{conditional shift} in a heterogeneous multi-task federated setting, where the relationship between observed inputs and segmentation targets varies across clients because each client corresponds to a different modality, anatomy, and imaging task. It also includes substantial \textbf{covariate shift} induced by site- and modality-specific data distributions, but the central challenge is the heterogeneity of the input--output mapping across clients.

In real-world federated medical imaging, institutions rarely share the same modality, anatomy, or segmentation task. This creates a more challenging setting than standard domain adaptation, because the model must learn under both privacy constraints and heterogeneous task definitions. \cite{shiranthika2026mucald} introduce \emph{MuCALD-SplitFed}, a causal--latent diffusion framework for privacy-preserving multi-task SplitFed medical image segmentation. The method combines causal representation learning, latent diffusion-based obfuscation, and domain-adversarial alignment to improve stability, reduce information leakage, and strengthen cross-task generalisation in heterogeneous clinical settings \cite{shiranthika2026mucald}.

\subsubsection{Experimental Dataset}
The framework is evaluated on five heterogeneous client datasets: a \emph{Blastocyst} dataset with 781 RGB embryo images \cite{lockhart2019multi}, \emph{HAM10K} with 10,015 dermatoscopic images \cite{tschandl2018ham10000}, an intrapartum transperineal ultrasound dataset (\emph{FHPsAOPMSB}) with 4,000 images \cite{lu2022jnu}, \emph{MosMed} with 2,800 lung CT images \cite{morozov2020mosmeddata}, and \emph{Kvasir-SEG} with 1,000 endoscopic polyp images \cite{jha2019kvasir}. Each client uses a fixed test set, with the remaining data split into training and validation partitions. This experimental design intentionally simulates a highly heterogeneous multi-client medical network with non-IID data, multiple modalities, and distinct segmentation targets \cite{shiranthika2026mucald}.

\subsubsection{Methodology}
MuCALD-SplitFed embeds causal modelling within the SplitFed latent space. The method first learns causal task representations and combines them with latent diffusion to obfuscate intermediate activations, thereby reducing privacy leakage. It then applies domain-adversarial alignment to discourage domain-specific leakage and improve cross-task robustness. The authors explicitly state that the framework is designed to disentangle task-relevant causal factors from domain-specific information, allowing the global model to remain useful despite client heterogeneity \cite{shiranthika2026mucald}.

\subsubsection{Result}
The paper reports that MuCALD-SplitFed consistently outperforms baseline SplitFed and state-of-the-art personalised and multi-task federated baselines across all five clients. In the main segmentation comparison, the method achieves an average Dice score of 0.816, compared with 0.394 for Baseline SplitFed and 0.626 for the strongest alternative reported in the table. The authors also report average IoU gains over communication rounds of 0.486 for U-Net, 0.416 for UNet3+, and 0.235 for SwinUNet. In addition to segmentation gains, the framework markedly degrades the reconstruction quality of transmitted feature maps, making reconstruction and membership-inference attacks substantially more difficult \cite{shiranthika2026mucald}.

\subsubsection{Clinical Relevance and Future Directions}
This work is clinically relevant because multi-institutional deployment increasingly requires privacy-preserving collaborative learning under heterogeneous data conditions. MuCALD-SplitFed shows how CTL-inspired representations can support both robustness and privacy in distributed medical imaging systems. For the present survey, it is especially valuable as an example of CTL moving beyond source--target transfer toward federated, multi-task, and deployment-oriented settings. Future work identified by the authors includes testing stronger adversarial models, scaling to larger client populations, and extending the framework to broader causal settings and additional modalities \cite{shiranthika2026mucald}.
%


\subsection{Causal Cross-Domain One-Shot Segmentation via Causal Intervention}

This case study addresses \textbf{covariate shift and domain confounding} in cross-domain medical image segmentation under an extreme data-scarcity setting, where only a single annotated example is available in the target domain. Unlike semi-supervised learning settings that rely on abundant unlabeled target data, one-shot cross-domain segmentation must explicitly handle both \textbf{distributional shift and severe label scarcity}, making causal intervention-based elimination of domain-related confounding factors a natural solution. Building on this motivation,\cite{hou2026eliminating} presents a causal inference–driven framework for eliminating domain-related confounding factors in cross-domain one-shot medical image segmentation. The key idea is to model domain-specific confounders that induce spurious correlations between imaging appearance and anatomical structure across different domains (e.g., hospitals, scanners, acquisition protocols). The method applies causal intervention to suppress these domain-induced biases and recover domain-invariant structural representations.

The proposed approach formulates cross-domain segmentation as a causal intervention problem, in which domain information serves as a confounder that influences both image appearance and segmentation predictions. By introducing intervention strategies within a causal graph formulation, the model reduces the effect of domain shift and enforces invariance of anatomical representations, enabling robust generalisation from a single labelled target sample.

\subsubsection{Experimental Datasets}
The framework is evaluated on three cross-domain benchmarks covering modality, sequence, and institutional shifts. The Abdominal MR$\leftrightarrow$CT dataset \cite{kavur2021chaos} focuses on multi-organ segmentation across imaging modalities, including liver, kidneys, and spleen. The Cardiac LGE$\leftrightarrow$bSSFP dataset \cite{zhuang2022cardiac} addresses cross-sequence cardiac segmentation involving LV blood pool, myocardium, and right ventricle structures. The Prostate UCLH$\leftrightarrow$NCI dataset \cite{li2023prototypical} evaluates cross-institution generalisation using multi-centre T2-weighted MRI scans for segmentation of bladder, central gland, and rectum.

\subsubsection{Methodology}
The proposed \textit{Domain Feature Correction Network} (DFCN) leverages causal inference to remove domain-specific confounding factors in cross-domain few-shot medical image segmentation through three complementary components. The Multi-Band Perturbation Rectification Module (MPRM) decomposes support and query features into frequency bands and applies perturbations followed by rectification to suppress non-causal domain variations and learn domain-invariant representations. The Domain Prompt Prototype Contrastive Learning (DPPCL) module incorporates learnable domain prompts within a contrastive framework to enhance intra-class prototype consistency and improve adaptation under limited supervision. The Prototype Relationship Optimisation (PRO) module further refines segmentation by modelling inter-region prototype dependencies to strengthen semantic coherence and structural consistency.

\subsubsection{Experimental Results}
DFCN achieves state-of-the-art performance across all evaluated cross-domain segmentation tasks as measured by DSC and IoU. The results highlight better feature alignment across domains, enhanced structural consistency, and reduced prediction uncertainty, confirming the effectiveness of causal feature correction in few-shot settings.

\subsubsection{Clinical Relevance and Future Directions}

This work is highly relevant for clinical deployment scenarios where acquiring annotations in new hospitals or imaging settings is extremely limited. By enabling one-shot cross-domain adaptation, the method reduces annotation burden and improves generalisation across institutions.

Future directions include extending the framework to other imaging modalities (e.g., ultrasound, PET), 3D volumetric segmentation, and multi-modal settings. Integrating causal intervention with semi-supervised or domain adaptation methods, as well as foundation models and multi-agent medical AI systems, could further improve robustness and scalability in real-world clinical applications.
            
        Having explored segmentation-focused solutions, the discussion shifts to classification, where causal learning principles enhance diagnostic accuracy despite limited training data and domain shifts.
            
    \subsection{Causal One-Shot MRI-Based Grading for Prostate Cancer}
         
        This case study addresses \textbf{covariate shift} between MRI scanner vendors (Siemens vs \ Philips) in a one-shot learning setting where labelled training data are extremely scarce.
        
        {Grading prostate cancer from MRI scans is essential for diagnosis and treatment, but challenges arise due to the limited labelled data and high variability in imaging across different MRI scanners and protocols. Domain shifts, particularly between scanners (e.g., Siemens vs Philips), further complicate model performance. One-shot learning, which trains models on minimal labelled data, offers a promising solution, but its effectiveness depends on the model’s ability to learn from a few examples. In this subsection, we will review a model proposed by \cite{carloni2023causality} that addresses this by integrating causal reasoning into the one-shot learning framework. Their model uses a Causality Extractor to identify and prioritise features causally linked to cancer grade, enabling accurate classification despite limited data and domain shifts.}
        
        \subsubsection{Experimental Dataset}
            {The PI-CAI prostate MRI dataset \cite{saha2023artificial}, consisting of 2,049 annotated T2-weighted MRI images, is used in this study. These images are lesion-annotated, highlighting cancerous regions in the prostate. To simulate domain shift, the dataset is split by MRI scanner vendor: training uses SIEMENS scanners, and validation and testing use Philips scanners \cite{carloni2023causality}. This split reflects real-world scenarios where models must generalise across different scanners. The dataset’s limited size further underscores the challenge of domain adaptation and makes it an ideal test case for one-shot learning.}
        
        \subsubsection{Methodology}
            {The proposed model, as shown in Figure \ref{fig:prostatearchitecture}, uses a simple convolutional backbone (ResNet18 \cite{targ2016resnet}) and a new causal mechanism. The ResNet18 backbone processes the input images and produces feature maps. The associated Causality Extractor in the model processes the feature maps and optimises pairwise conditional probabilities to produce a causality map. The causality map indicates asymmetric relationships among the feature maps. This collaborative method enables the model to distinguish causal features from effects. The causal factors of the map modify and augment the original feature maps within each feature vector space to prioritise causally important information. The modified feature map is concatenated with the original feature map, and the model makes its decision based on the combined causal information that aids classification. The entire model uses a one-shot meta-learning classification design for training. This method enables the model to learn from only a few examples and offers generalisability across imaging domains for prostate cancer classification. \cite{carloni2023causality}

            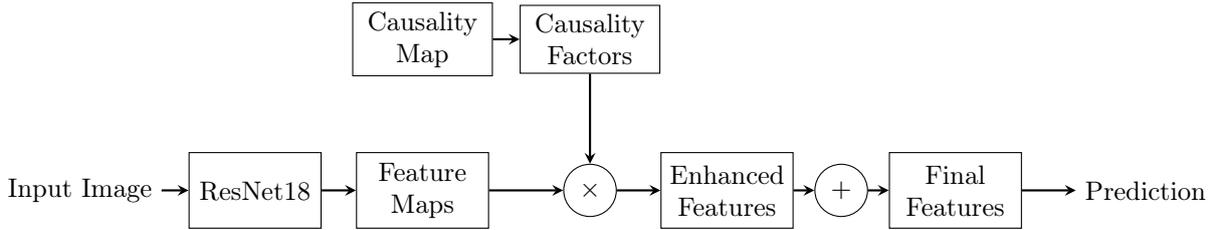
\begin{figure*}[h!]
    \centering
    \resizebox{\textwidth}{!}{%
    \begin{tikzpicture}[
        block/.style={draw, fill=white, rectangle, minimum height=1cm, minimum width=0.5cm, text width=1.5cm, align=center},
        smallblock/.style={draw, fill=white, rectangle, minimum height=0.8cm, minimum width=1.8cm, text width=1.6cm, align=center},
        arrow/.style={->, thick, >=stealth},
        label/.style={text width=2cm, align=center, font=\small}
    ]

    \node[] (input) at (0,0) {Input Image};

    \node[block] (resnet) at (2.3,0) {ResNet18};
    \node[block] (features) at (4.5,0) {Feature Maps};
    \node[smallblock] (causality) at (4.5,2) {Causality Map};
    \node[smallblock] (causalityFactors) at (6.7,2) {Causality Factors};
    \node[circle, draw, fill=white, minimum size=0.6cm] (multiply) at (6.7,0) {$\times$};
    \node[block] (enhanced) at (8.5,0) {Enhanced Features};
    \node[circle, draw, fill=white, minimum size=0.6cm] (concat) at (10,0) {$+$};
    \node[block] (final) at (11.5,0) {Final Features};
    \node[] (output) at (14,0) {Prediction};

    \draw[arrow] (input) -- (resnet);
    \draw[arrow] (resnet) -- (features);
    \draw[arrow] (causality) -- (causalityFactors);
    \draw[arrow] (causalityFactors)  -- (multiply);
    \draw[arrow] (features.east) -- (multiply.west);
    \draw[arrow] (multiply.east) -- (enhanced.west);
    \draw[arrow] (enhanced.east) -- (concat);
    \draw[arrow] (concat) -- (final);
    \draw[arrow] (final) -- (output);

    \end{tikzpicture}%
    }
    \caption{Causality-Driven ResNet18 for prostate cancer grading from MRI \cite{carloni2023causality}.}
    \label{fig:prostatearchitecture}
\end{figure*}

        \subsubsection{Result}
            {The model’s performance was evaluated on the PI-CAI dataset for four-class prostate cancer grading (ISUP 2–5). The results showed an Area Under the Receiver Operating Characteristic Curve (AUROC) of 0.614 for 4-class grading and 0.585 for improved focus on clinically relevant areas. The model outperformed the baseline by identifying features directly causally related to cancer grade \cite{carloni2023causality}. 
            }
        
        \subsubsection{Clinical Relevance and Future Directions}

        {The proposed causal one-shot learning framework offers key contributions: Cross-domain generalisation by learning domain-invariant features that reduce scanner-specific biases; few-shot learning that excels with limited labelled data, crucial for clinical settings with scarce annotations; clinical relevance, providing scalable solutions in environments where data privacy and scanner heterogeneity pose challenges; and future directions, including expanding to multi-modal tasks (e.g., MRI and CT integration), applying source-free domain adaptation, and integrating into clinical decision support systems for more efficient prostate cancer grading.}
        
        \subsection{Causality-Inspired Source-Free Domain Adaptation for Medical Image
Classification}
            This case study addresses \textbf{covariate and conditional shift} between chest X-ray acquisition devices and protocols. The clinical setting introduces an additional constraint: access to source-domain data is unavailable during adaptation due to patient privacy regulations, creating a source-free domain adaptation (SFDA) scenario where both the shift and the data access restriction must be handled simultaneously.

            {In traditional domain adaptation methods, the model requires access to source domain data during the adaptation phase, which raises both privacy concerns and storage costs. Given that medical images often contain sensitive patient data, this presents a significant challenge in clinical applications. In response, source-free domain adaptation (SFDA) has gained attention as a solution to this issue. SFDA utilises pre-trained models from the source domain, thereby eliminating the need for direct access to source-domain data during adaptation. The challenge, however, lies in effectively adapting to new target domains, which may differ significantly in acquisition conditions, scanners, or protocols. This section introduces a causality-inspired SFDA framework proposed by \cite{qiu2023causality} to improve the generalisation of medical image classification models and address domain shift issues.}
            
            \subsubsection{Experimental Dataset}
               {The Pulmonary Chest X-Ray Abnormalities dataset \cite{candemir2013lung, jaeger2013automatic} is used to evaluate the SFDA framework. The dataset includes two subsets, Montgomery and Shenzhen, which differ in imaging devices and acquisition conditions, thus simulating a domain shift \cite{qiu2023causality}. These differences make it a natural choice for evaluating domain-adaptation methods.}
                
            \subsubsection{Methodology}
                {The CSDA framework consists of two major components aimed at minimising domain shift by leveraging causal principles:}
                prototype-guided contrastive feature alignment and causality-driven interventions. Prototypes derived from the pre-trained source model produce class-level causal features to facilitate contrastive learning on target data, while minimising spurious correlations and acquisition bias through causal interventions through prototypes and augmentation \cite{qiu2023causality}.}

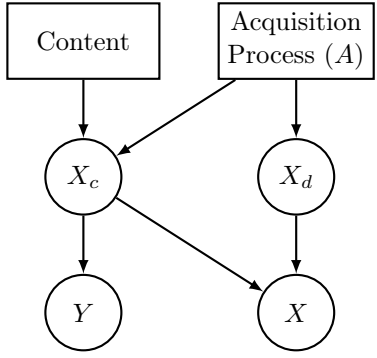
\begin{figure}[htbp]
                \centering
                \begin{tikzpicture}[->,>=latex, node distance=0.8cm, thick]

                    \node[draw, rectangle, minimum width=2cm, minimum height=1cm, fill=white] (Content) {Content};
                    \node[draw, rectangle, minimum width=2cm, minimum height=1cm, fill=white, right of=Content, xshift=2cm] (A) {\shortstack{Acquisition \\ Process ($A$)}};

                    \node[draw, circle, minimum size=1cm, fill=white, below of=Content, yshift=-1cm] (Xc) {$X_c$};
                    \node[draw, circle, minimum size=1cm, fill=white, below of=A, yshift=-1cm] (Xd) {$X_d$};

                    \node[draw, circle, minimum size=1cm, fill=white, below of=Xc, yshift=-1cm] (Y) {$Y$};
                    \node[draw, circle, minimum size=1cm, fill=white, below of=Xd, yshift=-1cm] (X) {$X$};

                    \draw[->] (Content) -- (Xc);
                    \draw[->] (A) -- (Xd);
                    \draw[->] (A) -- (Xc);
                    \draw[->] (Xc) -- (Y);
                    \draw[->] (Xc) -- (X);
                    \draw[->] (Xd) -- (X);

                \end{tikzpicture}
                \caption{Causal graph of medical image generation \cite{qiu2023causality}.}
                \label{fig:causal_a}
            \end{figure}

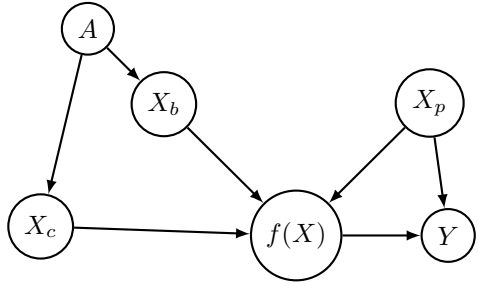
\begin{figure}[htbp]
                \centering
                \begin{tikzpicture}[->,>=latex, thick, node distance=1cm]

                \node[draw, circle] (A) {$A$};
                \node[draw, circle, below of=A, xshift=1cm] (Xb) {$X_b$};
                \node[draw, circle, below left=1cm and 1cm of Xb] (Xc) {$X_c$};
                \node[draw, circle, below right=1cm and 1cm of Xb] (Fx) {$f(X)$};
                \node[draw, circle, above right=1cm and 1cm of Fx] (Xp) {$X_p$};
                \node[draw, circle, right of=Fx, xshift=1cm] (Y) {$Y$};
                \draw[->] (A) -- (Xc);
                \draw[->] (A) -- (Xb);
                \draw[->] (Xb) -- (Fx);
                \draw[->] (Xc) -- (Fx);
                \draw[->] (Xp) -- (Fx);
                \draw[->] (Fx) -- (Y);
                \draw[->] (Xp) -- (Y);

                \end{tikzpicture}
                \caption{Causal graph of SFDA \cite{qiu2023causality}.}
            \label{fig:causal_b}
        \end{figure}

                {To further clarify the causal motivation of CSDA, Figs.~\ref{fig:causal_a} and \ref{fig:causal_b} illustrate the underlying mechanisms. As shown in Fig.~\ref{fig:causal_a}, medical images contain disease-relevant content features ($X_c$) and spurious background/domain features ($X_b$) introduced by the acquisition process ($A$). Although only $X_c$ causally determines the label $Y$, conventional models inadvertently exploit $X_b$, leading to a domain shift. In Fig.~\ref{fig:causal_b}, the SFDA setting introduces an additional confounder: source prototypes ($X_p$), which bias target feature learning. CSDA addresses these issues by applying augmentation-based interventions to weaken $X_b$ and backdoor adjustment to mitigate $X_p$, thus recovering the invariant mechanism $X_c \rightarrow Y$.}

            \subsubsection{Result}
                {The CSDA model was evaluated across two tasks in the Pulmonary Chest X-Ray Abnormalities dataset: the Montgomery → Shenzhen task, achieving an accuracy of 78.10\% and an AUC of 81.07\%, and the Shenzhen → Montgomery task, with an accuracy of 72.46\% and an AUC of 75.65\%. These results demonstrate the model's ability to adapt effectively to new imaging conditions, outperforming traditional domain adaptation methods that require access to source domain data \cite{qiu2023causality}.}
                
            \subsubsection{Clinical Relevance and Future Directions}
                {The CSDA framework offers several key contributions that enhance its clinical applicability. A major achievement of this work is its ability to maintain patient privacy, as the model can be deployed without accessing any source data. This becomes even more significant when class-specific prototypes are used, which guide alignment in the target domain, thereby strengthening the model's generalisation capacity in real-world medical contexts.}

    Building on advances in CTL for addressing domain shifts in medical image analysis, we now focus on counterfactual contrastive learning (CCL), which enhances robustness to acquisition-related variations.

    \subsection{Robust image representations with counterfactual contrastive learning}
            This case study addresses \textbf{covariate shift} attributable to scanner and acquisition hardware variability in chest radiography and mammography. The defining challenge is that different imaging devices produce systematically different image appearances even for the same patient, and standard contrastive learning methods learn to separate these device-specific signals rather than ignore them. Counterfactual image synthesis addresses this directly by simulating how each image would appear under a different scanner, enabling the model to learn representations that are invariant to acquisition hardware.
            Building on this motivation, \cite{roschewitz2025robust} introduces a framework designed to improve the robustness of contrastive learning in medical image analysis by generating counterfactual contrastive pairs. The authors propose using causal image synthesis techniques to simulate how images would appear under different acquisition conditions (e.g., different scanner types), enabling the model to focus on domain-invariant features and reducing bias introduced by variability in acquisition hardware.

            
            \subsubsection{Experimental Dataset}
                {The proposed framework was evaluated on five public medical image analysis datasets, including chest radiography and mammography, encompassing a variety of imaging hardware. These datasets, with a particular emphasis on different scanners and acquisition protocols, simulate real-world domain shifts. Specifically, the chest radiography evaluation used the PadChest \cite{bustos2020padchest} dataset, which includes data from two distinct scanners. In contrast, the mammography evaluation focused on the EMBED \cite{jeong2023emory} dataset, which comprises data from six scanners, with particular attention to scanners underrepresented in the dataset. 
                }
                
            \subsubsection{Methodology}
\begin{figure}[htbp]
                \centering
                \includegraphics[width=0.45\textwidth]{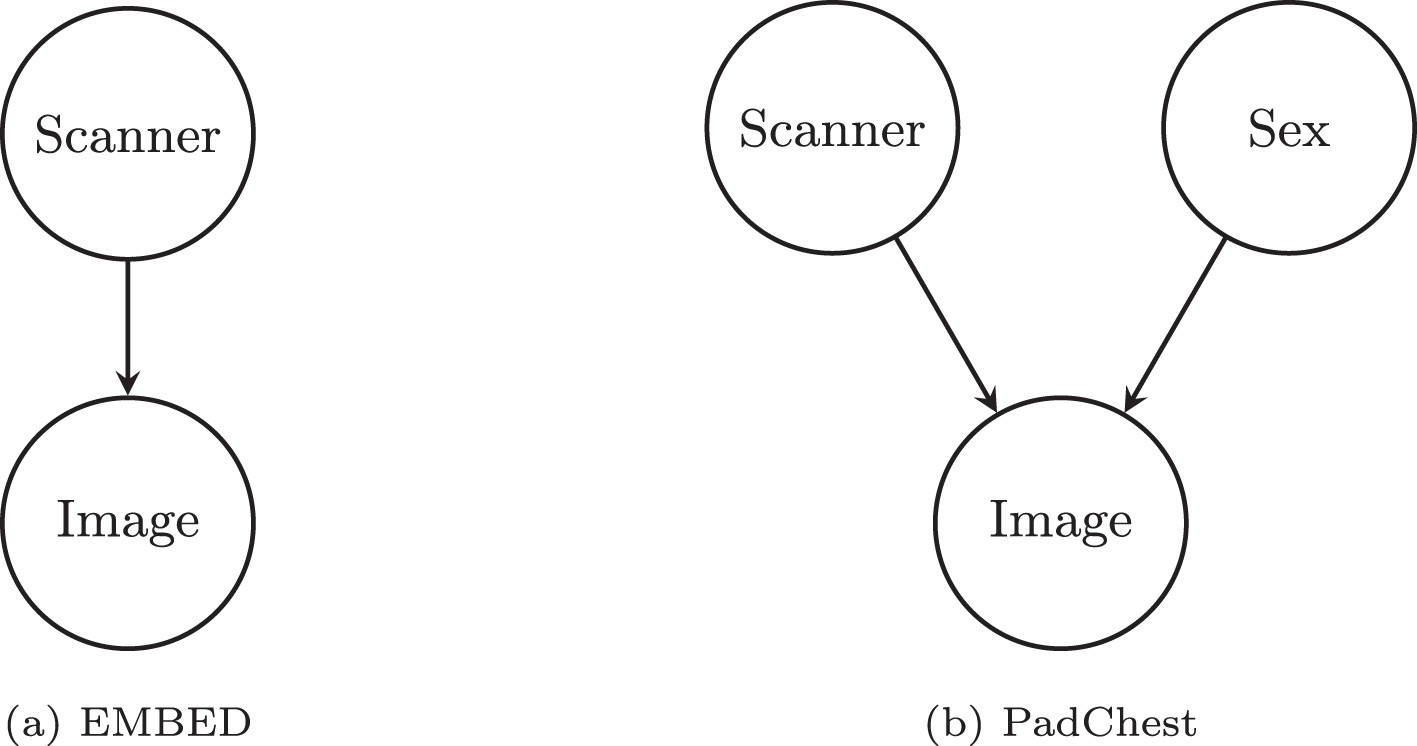}
                \caption{Causal graphs used to train the counterfactual image generation models \cite{roschewitz2409robust}.}
                \label{fig:RoschewitzFig}
            \end{figure}

                {The Counterfactual Contrastive Learning (CCL) framework addresses domain shifts in medical image analysis by combining causal image generation with contrastive learning.
                A visual representation of these causal graphs is provided in Figure \ref{fig:RoschewitzFig}. At its core, the model employs a Deep Structural Causal Model (DSCM) to generate counterfactual images simulating how images would appear if acquired with different scanners or protocols. This counterfactual generation is achieved through a Hierarchical Variational Autoencoder (HVAE), which produces realistic counterfactuals that preserve anatomical features while varying domain-specific attributes. In the contrastive learning phase, either the SimCLR or DINO-v2 objective is used, in which positive pairs are formed by pairing real images with their counterfactuals, enabling the model to learn domain-agnostic representations. These representations are then applied to downstream tasks, such as classification or segmentation, improving domain generalisation and robustness to acquisition shifts and subgroup disparities, especially in low-data scenarios. The model architecture incorporates a ResNet-50 encoder (for SimCLR) or a Vision Transformer (ViT for DINO-v2) for feature extraction, with counterfactuals generated by the HVAE and a contrastive learning objective that aligns real images with their counterfactuals. This architecture significantly enhances robustness and generalisation, particularly when training data is scarce and under-represented domains are present.}

            \subsubsection{Result}
                {The counterfactual contrastive learning framework outperformed traditional methods like SimCLR and DINO-v2, showing improved robustness to acquisition shifts, particularly for underrepresented scanners and limited labels. It also generalised well to external datasets, such as VinDR-Mammo and RSNA Pneumonia, and reduced disparities in subgroup performance, enhancing fairness across patient groups. t-SNE visualisations showed that CF-SimCLR minimised domain separation, focusing more on causal anatomical features rather than domain-specific variations, which is crucial for medical image analysis tasks.}
                
            \subsubsection{Clinical Relevance and Future Directions}
                {The clinical relevance of this work lies in its ability to enhance model robustness in clinical environments where domain shifts, to improve model robustness further, and acquisition protocols frequently occur. By incorporating counterfactual generation during training, the model can generalise more effectively across diverse imaging conditions, which is crucial for real-world medical image analysis. Furthermore, by utilising causal image synthesis, this method contributes to model fairness by addressing performance disparities across subgroups, such as gender-based differences. Future directions include extending this approach to other medical image analysis modalities, such as MRI and CT, and integrating it with multimodal image fusion (e.g., combining MRI with histopathological data) to improve model robustness further. Additionally, the framework could be applied to real-time clinical decision support systems, enabling models to quickly adapt to new imaging devices as they are introduced into clinical workflows.}

We now extend the exploration to accelerated MRI reconstruction, where causal reasoning techniques are employed to address domain shifts.


    
    \subsection{Illuminating the unseen: Advancing MRI domain generalisation through causality}
            This case study addresses \textbf{covariate and conditional shift} in accelerated MRI reconstruction, where differences in acquisition contrast, anatomical region, and undersampling pattern across scanners and protocols cause standard deep learning models to fail on unseen domains. Unlike the preceding case studies, the shift here operates at the level of the reconstruction mechanism itself, not only at the level of image appearance, making causal mechanism alignment rather than feature alignment the appropriate response. \cite{wang2025illuminating} presents GenCA-MRI, a domain generalisation (DG) framework designed specifically for deep learning-based accelerated MRI reconstruction. The framework develops a multi-level invariance approach -- image-level, feature-level, and mechanism-level -- to address the variability introduced by different acquisition strategies, contrasts, anatomical regions, and scanning conditions. Domain shifts often result in poor model performance when exposed to unseen data. To overcome this, the paper develops a multi-level invariance framework: image-level, feature-level, and mechanism-level invariance. The most novel aspect is GenCA-MRI, a causal mechanism alignment approach that aligns intrinsic causal relationships across domains, ensuring consistent performance on new, previously unseen datasets.

            
            \subsubsection{Experimental Dataset}
                {The framework was evaluated on two public MRI datasets: fastMRI \cite{knoll2020fastmri} and IXI (http://brain-development.org/ixi-dataset/), which contain various contrasts and anatomical regions. These datasets simulate common domain shifts in medical image analysis, such as variations in acquisition protocols and scanning conditions across scanners or patients. The evaluation was conducted across multiple settings, including different acceleration factors (2$\times$ to 8$\times$), to assess the robustness of the proposed framework to challenges such as contrast, anatomical regions, and unseen undersampling patterns.
                }
                
            \subsubsection{Methodology}
                {The paper employs a causal transfer learning approach, which is a form of domain generalisation. Unlike traditional domain-adaptation methods that require access to target-domain data, this framework avoids such dependence by leveraging causal reasoning to ensure domain-invariant representations. The approach comprises the following steps: (i) Image-Level Fidelity Consistency: The model ensures that reconstructed images maintain high quality across domains by using adversarial loss to constrain image fidelity; (ii) Feature-Level Invariance: Feature alignment techniques are employed to align hidden representations from different domains, ensuring that critical features (e.g., edges, textures) remain consistent; and (iii) Mechanism-Level Invariance (GenCA-MRI): The paper’s most significant contribution, GenCA-MRI, aligns the causal mechanisms underlying MRI reconstruction quality across domains. This is achieved by quantifying causal relationships with an Average-Causal-Effect (ACE) module and applying a causal alignment loss to ensure consistent imaging mechanisms across different acquisition strategies.}

            \subsubsection{Result}
                {The experimental results demonstrate that GenCA-MRI outperforms existing methods in multiple metrics, including PSNR, SSIM, and MSE. It achieves significant improvements in reconstruction quality, particularly under challenging domain shifts, such as unseen contrasts or anatomical regions. The framework’s performance was evaluated at various acceleration factors (2$\times$, 4$\times$, 6$\times$ and 8$\times$) and consistently showed superior results. Notably, GenCA-MRI demonstrated an improvement in PSNR of up to 2.15 dB on the fastMRI dataset \cite{knoll2020fastmri} and 1.24 dB on the IXI dataset at 8$\times$ acceleration, highlighting its robustness and adaptability across different MRI protocols.}
                
            \subsubsection{Clinical Relevance and Future Directions}
                {The clinical relevance of GenCA-MRI lies in its ability to generalise across different MRI datasets without the need for target domain data, making it particularly suitable for environments where access to diverse clinical data is limited or constrained due to privacy concerns. This is essential for real-world clinical applications, where new MRI machines and protocols are constantly introduced. The ability to maintain high reconstruction quality, even in the face of domain shifts, directly impacts diagnostic accuracy and efficiency in clinical workflows. Future directions for GenCA-MRI include expanding its applications to multimodal and multidomain tasks, such as integrating CT and PET imaging for comprehensive diagnostic systems. Additionally, enhancing the framework for source-free domain adaptation could enable real-time clinical applications where access to source-domain data is limited. Integrating GenCA-MRI into clinical decision support systems could further streamline diagnostic workflows, improving efficiency and accuracy across diverse healthcare settings. These advancements would make GenCA-MRI a versatile and powerful tool for real-world medical image analysis.}

{These studies demonstrate that CTL provides a unified framework to address data scarcity, domain shift, and spurious correlations in medical image analysis. Causal mechanisms, such as SCMs, interventions, and invariance, have been shown to improve generalisation across segmentation and classification tasks. The applicability of CTL extends across multiple imaging modalities, including fundus, MRI, CT, histopathology, and chest X-ray, where it supports robust transfer and domain-invariant learning (see Table~\ref{applications_modalities}). Beyond its practical benefits, CTL improves not only accuracy but also fairness, privacy, and robustness, making it a compelling paradigm for deployment in real-world clinical settings.}

\begin{table*}[h]
\centering
\scriptsize
\renewcommand{\arraystretch}{1.2}

\caption{Applications of Causal Transfer Learning across Medical Imaging Modalities}
\label{applications_modalities}

\begin{tabular}{@{}>{\raggedright\arraybackslash}p{4cm}
                >{\raggedright\arraybackslash}p{11cm}@{}}
\toprule

\textbf{Modality / Task} &
\textbf{CTL Method and Benefit} \\
\midrule

\multicolumn{2}{l}{\textbf{(A) Domain adaptation and generalisation}} \\
\midrule

Fundus Imaging \cite{li2025causal} \newline
Cup-to-disc ratio estimation for glaucoma diagnosis
&
Domain adaptation and structural causal models were used to improve generalisation across imaging devices with different styles.
\\

CT and MRI \cite{chen2025generalizable} \newline
Single-source segmentation (cross-modality)
&
Structural Causal Models (SCMs), combined with diffusion-based style interventions, enabled the simulation of missing modalities and reliable segmentation across unseen modalities.
\\

Multi-domain datasets \cite{ouyang2022causality} \newline
Single-source domain generalisation for segmentation
&
Causality-inspired augmentations (GIN + IPA) improved cross-domain segmentation performance by focusing on domain-invariant anatomical structures.
\\

Chest X-ray \cite{qiu2023causality} \newline
Classification under domain shift
&
Causality-inspired source-free domain adaptation using prototype-guided contrastive feature alignment and causal interventions reduced performance degradation across hospitals and imaging devices.
\\

Chest X-ray (PadChest) \cite{bustos2020padchest} and Mammography (EMBED) \cite{jeong2023emory} \newline
Classification under domain shift
&
Counterfactual Contrastive Learning (CCL) improved robustness to acquisition shifts, reduced bias, and enhanced generalisation for under-represented scanners.
\\

\midrule
\multicolumn{2}{l}{\textbf{(B) Segmentation and reconstruction}} \\
\midrule

Cardiac MRI (ACDC) \cite{bernard2018deep}, Pancreas-CT \cite{clark2013cancer}, and Brain Tumor MRI (BraTS'19) \cite{bakas2017advancing} \newline
Segmentation of cardiac, pancreatic, and tumour structures
&
Causality-inspired Semi-supervised Learning (CauSSL) improved segmentation with limited labelled data by ensuring network independence and enhancing efficiency across both 2D and 3D architectures.
\\

MRI (fastMRI) \cite{knoll2020fastmri} and MRI (IXI) \newline
Accelerated MRI reconstruction
&
GenCA-MRI (Causal Mechanism Alignment) improved reconstruction across domain shifts and unseen contrasts by aligning causal relationships and increasing robustness to varying MRI protocols.
\\

Abdominal (MR$\leftrightarrow$CT), Cardiac, and Prostate \cite{hou2026eliminating} 
& 
Domain Feature Correction Network (DFCN): Employs causal intervention (MPRM) and prototype optimisation (PRO) to remove domain-specific confounders, enabling robust segmentation from a single target-domain label 
\\

\midrule
\multicolumn{2}{l}{\textbf{(C) Interpretability and causal representation}} \\
\midrule

Histopathology \cite{carloni2023causality} \newline
Cancer subtype classification
&
A causal model integrated with transfer learning enabled the identification of causal features beyond simple correlations, improving interpretability.
\\

\bottomrule
\end{tabular}

\end{table*}

\newpage
To better understand how CTL methods achieve these benefits across various imaging tasks, a consolidated mapping of the reviewed methods by task, shift type, and causal assumption is provided in Table~\ref{tab:ctl_methods_mapping}. This table synthesises the representative families of methods discussed in Sections~5 and~6, illustrating how different CTL approaches address challenges such as domain/environment shifts, covariate and label shifts, and more, while also highlighting the underlying causal assumptions that drive their effectiveness.

\begin{table*}[h]
\centering
\scriptsize
\caption{Mapping of CTL methods reviewed in Sections~5 and~6, organised by imaging task, domain/shift type, and underlying causal assumption.}
\label{tab:ctl_methods_mapping}
\begin{tabular}{@{}>{\raggedright\arraybackslash}p{3.5cm}
                >{\raggedright\arraybackslash}p{3.2cm}
                >{\raggedright\arraybackslash}p{3.8cm}
                >{\raggedright\arraybackslash}p{4.2cm}@{}}
\toprule
\textbf{Method Family} &
\textbf{Imaging Task(s)} &
\textbf{Shift Type Addressed} &
\textbf{Causal Assumption} \\
\midrule

Invariant Risk Minimisation (IRM), REx \cite{arjovsky2020invariantriskminimization}
& Classification
& Domain / environment shift
& Existence of invariant causal predictors across environments \\
\midrule

Causal representation learning \cite{scholkopf2021toward, ahuja2023interventional}
& Classification, segmentation
& Covariate and mechanism shift
& Latent SCM governing representations \\
\midrule

Disentangled causal factor models \cite{yang2021causalvae, suter2019robustly}
& Reconstruction, multimodal imaging
& Covariate shift
& Independent causal generative factors \\
\midrule

Domain-adversarial learning (causality-inspired) \cite{lv2022causality}
& Classification, segmentation
& Covariate / scanner shift
& Anti-causal setting with invariant conditional mechanisms \\
\midrule

Conditional domain adaptation \cite{long2018conditional, zhang2013domain, gong2016domain}
& Classification
& Label shift
& Invariance of $P(Y \mid X)$ across domains \\
\midrule

Causal feature selection \cite{guyon2007causal}
& Classification
& Spurious correlation shift
& Selected features correspond to causal parents of the label \\
\midrule

Interventional data augmentation \cite{ilse2021selecting}
& Classification, segmentation
& Interventional / environment shift
& Known or assumed intervention targets in a causal graph \\
\midrule

Style--content separation models \cite{zhang2018separating}
& Segmentation, reconstruction
& Scanner / acquisition shift
& Style variables are non-causal nuisance factors \\
\midrule

Counterfactual data generation \cite{chang2021towards}
& Classification
& Concept shift
& Explicit SCM enabling counterfactual reasoning \\
\midrule

SCM-based domain adaptation
& Classification
& Environment and population shift
& Explicit causal graph with back-door adjustment \\
\midrule

Causal regularisation losses \cite{janzing2019causal, bahadori2017causal}
& Classification
& Covariate and label shift
& Penalisation of non-causal dependencies \\
\midrule

Meta-learning for causal generalisation \cite{wang2022meta, ton2021meta}
& Classification
& Domain shift
& Stable causal mechanisms across training tasks \\
\midrule

Multi-site harmonisation (e.g.\ ComBat-inspired) \cite{gardner2025combatls}
& Reconstruction, segmentation
& Site / scanner shift
& Additive causal effects of site-specific factors \\
\midrule

Longitudinal causal modelling \cite{arjas2004causal, arkhangelsky2024causal}
& Longitudinal imaging
& Temporal distribution shift
& Time-dependent SCM with causal transitions \\
\midrule

Multimodal causal fusion \cite{wu2005multimodal, wu2025multimodal}
& Multimodal imaging
& Modality shift
& Causal ordering and interaction between modalities \\
\midrule

Causality-aware anomaly detection \cite{xiao2025causality}
& Anomaly detection
& Distribution shift
& Anomalies violate learned causal mechanisms \\
\midrule

Causal Feature Correction (DFCN) \cite{hou2026eliminating}
& Cross-domain one-shot segmentation
& Covariate shift and domain confounding
& Causal intervention to eliminate non-causal domain variations via frequency-band perturbation \\
\bottomrule

\end{tabular}
\end{table*}

\section{Challenges and Limitations}

While CTL holds great promise in medical image analysis, numerous challenges and limitations must be overcome before the technology is widely accepted in clinical practice. This section explains the main challenges that researchers and clinicians face in implementing CTL (see Table \ref{Tablechallanges}).

\begin{table*}[!t]
    \centering
    \scriptsize
    \caption{Challenges in Causal Transfer Learning}
    \begin{tabular}{@{}>{\raggedright}p{4cm} p{5cm} p{5cm}@{}}
        \toprule
        \textbf{Challenge} & \textbf{Description} & \textbf{Potential Solutions} \\
        \midrule
        Causal Identifiability Across Clinical Domains & Determining whether invariant causal mechanisms remain identifiable under evolving clinical distributions and imaging protocols.& Development of causal stability theory, domain-aware SCMs, and uncertainty-aware causal discovery methods. \\ 
        \midrule
        Evaluation of Causal Generalisation & Existing metrics do not directly measure causal validity, intervention robustness, or counterfactual consistency.& Designing benchmark datasets and evaluation protocols based on interventions, counterfactual reasoning, and causal stress testing. \\ 
        \midrule
        Trade-off Between Invariance and Clinical Specificity & Excessive invariance constraints may suppress clinically meaningful demographic or modality-specific information.& Adaptive invariance frameworks and context-aware causal regularisation methods. \\ 
        \midrule
        Learning Causal Representations from High-Dimensional Imaging Data & High-dimensional medical imaging data (e.g., MRI, CT) contain complex patterns, making it difficult to learn accurate causal representations.& The development of advanced causal models, such as causal convolutional neural networks and graph-based causal inference, can help identify and learn the underlying causal structure from imaging data. \\ 
        \midrule
        Causal Discovery & Identifying and inferring the correct causal relationships in imaging data, where features are correlated but not causally linked. & Implementing causal discovery algorithms, such as constraint-based methods (e.g., the PC algorithm) and causal Bayesian networks, can help uncover valid causal structures. \cite{pearl2009causality} \\
        \midrule
        Defining Causal Structure for Images & Developing a formal framework for structuring causal relationships within imaging data to make the models more interpretable and reliable. & Employing causal graphical models and counterfactual reasoning frameworks (e.g., CausalGAN) to provide a robust framework for defining causal relationships in images. \\
        \midrule
        Limited Benchmark Datasets & Availability of a few benchmark datasets that incorporate causal labels or counterfactual information, making it difficult to evaluate CTL models effectively. & Creating and curating causal benchmark datasets for medical image analysis that include annotated causal information or counterfactual scenarios. Initiatives like CaDiRa can help address these gaps \cite{zhu2025counterfactual}. \\
        \midrule
        Scalability & Difficulty in deploying models that generalise across diverse clinical settings due to computational constraints. & Development of more efficient algorithms and model architectures, such as multi-scale causal learning frameworks and distributed causal models. \\
        \midrule
        Clinical Validation & Lack of rigorous validation studies to confirm the model's effectiveness in real-world settings, undermining clinicians' trust. & Conducting clinical validation studies, including retrospective and multi-institutional evaluations, to assess model performance in diverse clinical environments. \\
        \midrule
        Ethical Considerations & Risks of biased outcomes and disparities arising from data selection and model training processes. & Implementing bias detection mechanisms and ethical guidelines for model development, including transparency in model decision-making. \\
        \midrule
        Interpretability & Challenges in explaining model decisions and causal relationships to clinicians, hindering adoption. & Developing interpretable models and visualisation tools for clear communication of results, such as causal saliency maps for medical image analysis. \\
        \bottomrule
    \end{tabular}
    \label{Tablechallanges}
\end{table*}

\subsection{Scalability}

One of the significant challenges in CTL is scalability, which refers to the ability to deploy models that can be easily generalised across different clinical settings. As medical datasets grow in size and complexity, there is an increasing challenge in sustaining the efficiency and interpretability of causal models.

The high computational requirements of causal discovery and counterfactual inference algorithms often make them unsuitable for real-time applications. The above situation clearly calls for the development of new algorithms that are more efficient and architectures that support these processes.

The other cardinal aspect of this challenge is high dimensionality. Most medical image analysis datasets include multiple imaging modalities, anatomical variation, and varying spatial resolutions. It is very challenging to address this complexity while respecting causal assumptions. Most existing causal inference methods, particularly SCM-based approaches, require substantial computational resources, particularly when applied to large datasets.

Therefore, there is a strong need for efficient algorithms to perform causal inference without bearing high computational costs. In clinical settings, timely decisions are often required. Thus, CTL models must be not only accurate but also sufficiently fast to support clinicians in real-time decision-making.

\subsection{Clinical Validation}

The biggest challenge in the development of CTL models is that of clinical validation. While these may perform well in the carefully controlled environments of research studies, their performance in real-world settings remains largely untested. The complexity of health care, including patient demographics, imaging protocols, and clinician practices, can significantly affect a model's performance in practice.

It may well be that results like these are not generalisable to all patient populations, given variations in patient demographics, health conditions, and treatment approaches. It requires extensive validation studies involving diverse patient groups and clinical settings to determine whether the CTL models can be generalised.


Another important consideration is the temporal robustness of CTL models. Model performance should be evaluated longitudinally, as patient responses and treatment outcomes may evolve over time. Longitudinal studies are therefore necessary to ensure that CTL models remain robust and effective across different stages of disease progression and treatment.


\subsection{Ethical Considerations}

The integration of CTL into medical image analysis raises ethical issues that must be addressed. One primary concern is that outcomes are biased by imbalances in the training data or by incorrect assumptions about causal relations. Such biases can also amplify currently existing disparities in patient care; therefore, strategies must be implemented to identify and mitigate bias during both model training and validation.

However, most causal models are complex and opaque, which undermines transparency and trust in clinical decision-making. Models can also perpetuate existing healthcare disparities if they are trained on biased datasets; thus, fairness and equity must be well considered in the data.

It also raises important questions regarding consent and privacy in the use of patient data to develop these models. Therefore, the creation of an ethical framework governing the responsible and transparent handling of patient data by these systems is essential.

It would also be relevant to consider the impact of CTL models on clinician judgment and patient interactions. Increased reliance on algorithmic decision-making may weaken the clinician-patient relationship and ultimately shake the public's trust in medical professionals. Thus, finding an appropriate balance between the use of advanced AI models and transparent oversight in clinical practice is instrumental in delivering effective and ethical healthcare.

\subsection{Interpretability and Explainability}

Causality and explainable artificial intelligence (XAI) go hand in hand; both aim to help us understand how AI models make decisions. On the other hand, causality concerns how one variable causes another, whereas XAI focuses on making the outputs of AI models transparent and interpretable. When we incorporate causality into XAI, we can generate explanations that precisely explain how specific inputs lead to certain outcomes. This would make the AI systems more transparent and trustworthy, as one can see the reasons for a model's prediction. Causality is not identical to XAI, although it plays a pivotal role in improving the understanding and explanation of AI decisions.

As such, presenting these causal relations in a simplified manner is necessary without sacrificing model accuracy. Similarly, develop user-friendly tools and interfaces that present causal insights visually to increase clinician engagement and support better decision-making. It may also require promoting training programs to help clinicians confidently interpret such insights. More importantly, feedback loops, in which clinicians can comment on model outputs, enhance interpretability and adaptability in CTL models. Collaborative approaches might lead to more personalised and effective care solutions.

\section{Current Research Directions}

The field of CTL in medical image analysis offers numerous avenues for future research. This section outlines several promising directions for addressing existing challenges and enhancing the impact of CTL.

\subsection{Advanced Causal Discovery Techniques}

Future research should therefore strive to develop causal discovery methods tailored to estimating causal relations in high-dimensional medical image analysis data. More innovative methods, building on current state-of-the-art techniques such as Bayesian networks, graphical models, and deep learning, can move us closer to understanding complex causal structures. Causal reasoning will be easily incorporated into CTL frameworks by improving these approaches. This will be of great importance in improving the accuracy of predictions and personalised treatment recommendations, thereby improving patient outcomes in clinical settings.

\subsection{Integration with Multimodal Data}

The integration of CTL with multimodal data, including EHRs, genomic information, and multiple imaging modalities, has created a deep research line in medical image analysis. Such integration of diverse data types enables the development of holistic models that capture multiple causal factors affecting patient outcomes. For example, integrating imaging data with clinical and genomic information enables a holistic understanding of disease; models can now account for a broader range of factors influencing patient health. The multimodal approach can thus enhance decision-making frameworks in clinical practice and, in turn, improve diagnostic accuracy and inform the appropriate therapies for each patient.

\subsection{Causality Aware Generative Modelling}
A promising direction for integrating causal reasoning with generative models in medical image analysis is causality-aware generative modelling. These models explicitly integrate causal inference principles to enable controllable image generation and counterfactual simulation. By using such models, we can generate synthetic training data reflecting causal interventions, which is particularly valuable for counterfactual data augmentation. This approach enhances the diversity and relevance of training datasets, improving model robustness and generalisability across clinical scenarios. Recent work has explored the use of causal generative models for medical image synthesis and counterfactual image generation, thereby enabling more accurate simulations of clinical outcomes under various hypothetical conditions \cite{vigneshwaran2024macaw, ibrahim2024semi}.

\subsection{Medical Imaging Enhancement via CTL}

Although DL-based reconstruction and enhancement for undersampled low-field MRI have been demonstrated, including direct super-resolution and image quality enhancement from undersampled k-space data \cite{ayde2022deep, anyimadu2026direct}, transfer-learning strategies for accelerated or reconstructed MRI across field strengths, anatomies, and sampling patterns have also proven effective \cite{arshad2021transfer, dar2020transfer, lv2021transfer, bi2023linear}. Recent work has further highlighted the importance of out-of-distribution generalisation in low-field MRI enhancement under undersampled acquisition settings \cite{anyimadu2026low}. However, we found a paucity of published work that combines undersampled low-field MRI reconstruction or enhancement with an explicit CTL framework. To address this, future research can focus on applying causal inference techniques, such as causal mechanism alignment \cite{wang2025illuminating}, to improve generalisation, reduce bias, and enhance robustness to domain shifts in low-field MRI, ultimately advancing undersampled low-field MRI reconstruction and enhancement.

\subsection{Addressing Ethical and Equity Concerns}

Ethical and equity considerations must be addressed as CTL models are developed and deployed. Future research should focus on frameworks for assessing fairness in CTL applications and on methods to reduce bias in training datasets. Collaboration among researchers, clinicians, and ethicists is vital to ensure that CTL models increase access to and equity in health care. Developers are constrained to create effective CTL models based on sound scientific principles. The clinicians' insights indicate how these models can be applied in real-world medical settings to meet the pragmatic needs of both patients and health care providers. Ethicists help determine the most effective ways to examine the ethical implications of these technologies, to avoid harm to vulnerable populations and ensure that beneficence in CTL is appropriately shared. By collaborating, these organisations can help ensure that the CTL models under development make healthcare more accessible to all and do not inadvertently widen existing inequitable access gaps.

\subsection{Clinical Implementation}

Future research in this area should increasingly focus on implementation studies in real-world settings that test the applicability and effectiveness of these models across diverse healthcare settings. Involving stakeholders, including health professionals and patients, through rigorous evaluation of their experiences with the usability, acceptability, and adaptability of CTL applications will also substantially advance the translation of theoretical progress into practical benefits for patient care.

\section{Healthcare Datasets for CTL}
CTL shows great potential in addressing domain shift and data scarcity, and in enhancing model generalisation in medical image analysis. However, not all healthcare datasets are suitable for CTL. Identifying datasets that offer diverse modalities, cross-institutional variability, and large sample sizes is essential for effective model training. This section highlights healthcare datasets well-suited for CTL; see Table \ref{datasets_ctl} for a comprehensive list. Specifically, several healthcare datasets are well-suited to CTL, as they naturally exhibit domain shifts arising from variations in imaging devices, acquisition protocols, and clinical environments. In ophthalmic imaging, datasets such as REFUGE \cite{orlando2020refuge}, DRISHTI-GS \cite{sivaswamy2015comprehensive}, and RIM-ONE-r3-all \cite{fumero2011rim} contain fundus images acquired using different camera systems, lighting conditions, and resolutions. These factors introduce substantial domain shifts that hinder model generalisation. CTL has been shown to effectively mitigate such shifts, with \cite{li2025causal} demonstrating significant improvements in cross-domain fundus image segmentation performance.

Similar challenges arise in cross-modality medical image segmentation. In abdominal segmentation tasks, CT and MRI datasets from MICCAI 2015 \cite{landman2015miccai}, AMOS \cite{ji2022amos}, and CHAOS \cite{kavur2021chaos} present substantial modality-induced variability, particularly when segmenting organs such as the liver, kidneys, and spleen. CTL enables effective generalisation across these modalities by disentangling causal anatomical features from modality-specific artefacts. Related approaches have been applied to lumbar spine segmentation, combining CT data from \cite{sekuboyina2020labeling}, T2-weighted MRI from \cite{pang2020spineparsenet}, and X-ray images from \cite{klinwichit2023buu} to achieve robust vertebrae segmentation across heterogeneous imaging domains. Lung segmentation further illustrates the benefits of CTL, with \cite{chen2025generalizable} leveraging CT and X-ray data from \cite{yang2017data} and \cite{danilov2022automatic} to improve segmentation robustness under cross-domain shifts.

Domain shifts are also prevalent within MRI data due to variations in imaging sequences and acquisition protocols. The Cardiac Cross-sequence dataset \cite{zhuang2022cardiac}, which involves adaptation from balanced steady-state free precession (bSSFP) MRI to late gadolinium enhancement (LGE) MRI, poses a challenging cross-sequence segmentation problem. CTL has been shown to enhance robustness to such sequence variations, facilitating effective cross-sequence adaptation. Similarly, the Prostate Cross-site dataset \cite{liu2020shape,bloch2015nci,lemaitre2015computer,litjens2014evaluation} aggregates prostate MRI scans from six different clinical sites, introducing pronounced cross-site variability. CTL has been successfully applied to this setting to improve consistency across acquisition protocols, as demonstrated by \cite{ouyang2022causality}. Larger-scale datasets such as PI-CAI \cite{saha2023artificial}, which includes over 10,000 prostate MRI examinations collected across multiple European centres, further highlight the relevance of CTL for addressing cross-centre and cross-device variability. This dataset has been used by \cite{carloni2023causality} to support one-shot learning in prostate cancer grading.

Chest radiography datasets provide additional examples of acquisition-induced domain shifts. The Montgomery and Shenzhen datasets \cite{jaeger2014two}, commonly used for pulmonary disease detection, differ substantially in imaging protocols and scanner characteristics. CTL methods, including source-free domain adaptation and causal feature alignment, have been shown to improve cross-dataset performance in this setting significantly \cite{qiu2023causality}. The PadChest dataset \cite{bustos2020padchest}, which contains chest X-ray images acquired from two distinct scanners, further supports domain counterfactual analysis in tasks such as pneumonia detection. Models trained on PadChest are often evaluated on external datasets, including RSNA Pneumonia Detection \cite{stein2018rsna,shih2019augmenting} and CheXpert \cite{irvin2019chexpert}, demonstrating their suitability for assessing generalisation to unseen scanner domains.

In mammography, large-scale datasets such as EMBED \cite{jeong2023emory} include over 300,000 scans acquired from six different imaging devices, resulting in substantial scanner-induced variability. Underrepresented devices, such as the Selenia Dimensions scanner, pose a particularly challenging domain-adaptation scenario that benefits from CTL-based approaches. The VinDR-Mammo dataset \cite{nguyen2023vindr}, collected in Vietnam, further expands this setting by introducing domain shifts across distinct acquisition environments. These datasets have been used by \cite{roschewitz2409robust} to learn robust image representations through counterfactual contrastive learning, yielding improved generalisation across mammography domains.

Several widely used segmentation benchmarks are also well suited to CTL due to their multimodal nature. These include ACDC \cite{bernard2018deep} for cardiac segmentation, Pancreas-CT \cite{clark2013cancer,roth2016data,roth2015deeporgan} for abdominal organ segmentation, and BraTS'19 \cite{bakas2017advancing,bakas2018identifying,baid2021rsna,menze2014multimodal} for brain tumour segmentation. The substantial variability across imaging modalities and acquisition settings in these datasets makes them valuable testbeds for evaluating causal generalisation, as explored by \cite{miao2023caussl}.

Finally, datasets such as IXI\footnote{\url{http://brain-development.org/ixi-dataset/}} and fastMRI \cite{knoll2020fastmri} are particularly relevant for CTL in MRI reconstruction and representation learning. IXI provides multi-contrast MRI data across T1-, T2-, PD-, and FLAIR-weighted images, while fastMRI includes large-scale multi-coil \textit{k}-space data from knee, brain, and prostate MRI acquisitions. These datasets introduce domain shifts across anatomical regions, imaging protocols, and undersampling patterns, making them suitable for evaluating CTL’s ability to learn domain-invariant representations. Recent work by \cite{wang2025illuminating} demonstrates the effectiveness of CTL in improving robustness and generalisation in such reconstruction settings.

\begin{table*}[h]
    \centering
    \scriptsize
    \caption{Datasets Suited for CTL}
    \label{datasets_ctl}
    \begin{tabular}{@{}>{\raggedright\arraybackslash}p{5cm} >{\raggedright\arraybackslash}p{9cm}@{}}
        \toprule
        \textbf{Dataset/Task} & \textbf{Relevance to CTL} \\
        \midrule
        \href{https://refuge.grand-challenge.org/}{REFUGE} \cite{orlando2020refuge}, 
        \href{http://cvit.iiit.ac.in/projects/mip/drishti-gs/mip-dataset2/Home.php}{DRISHTI-GS} \cite{sivaswamy2015comprehensive}, 
        RIM-ONE-r3-all \cite{fumero2011rim} 
        & Fundus imaging for domain adaptation in glaucoma diagnosis across camera systems and lighting conditions. \\
        \midrule

        Abdominal Segmentation (AS) \cite{landman2015miccai}, \cite{ji2022amos}, \cite{kavur2021chaos} 
        & CT and MRI datasets for cross-modality abdominal organ segmentation, such as liver and kidneys. \\
        \midrule

        Lumbar Spine Segmentation (LSS) \cite{sekuboyina2020labeling}, \cite{pang2020spineparsenet}, \cite{klinwichit2023buu} 
        & Cross-modality segmentation of vertebrae using CT, MRI, and X-ray data for generalisation across imaging modalities. \\
        \midrule

        Lung Segmentation (LS) \cite{yang2017data}, \cite{danilov2022automatic} 
        & CT and X-ray datasets for segmenting left and right lungs across varying imaging conditions. \\
        \midrule

        Cardiac Cross-sequence \cite{zhuang2022cardiac} 
        & Cardiac MRI data transitioning from bSSFP to LGE MRI for sequence adaptation. \\
        \midrule

        Prostate Cross-site \cite{liu2020shape, bloch2015nci, lemaitre2015computer, litjens2014evaluation} 
        & Prostate MRI data from multiple sites to address cross-site variability. \\
        \midrule

        PI-CAI \cite{saha2023artificial} 
        & Prostate MRI data for cross-centre variation handling and one-shot learning. \\
        \midrule

        Chest X-ray (Montgomery \cite{jaeger2014two}, Shenzhen) 
        & Chest X-ray data for pulmonary disease detection under domain shift using CTL techniques. \\
        \midrule

        \href{https://bustos2020padchest.com/}{PadChest} \cite{bustos2020padchest}, 
        \href{https://www.rsna.org/rsna-pneumonia-detection-challenge}{RSNA Pneumonia Detection} \cite{stein2018rsna}, 
        \href{https://stanfordmlgroup.github.io/chexpert/}{CheXpert} \cite{irvin2019chexpert} 
        & Chest radiographs from different scanners and acquisition protocols; CTL improves model generalisation across scanning setups. \\
        \midrule

        EMBED \cite{jeong2023emory}, VinDR-Mammo \cite{nguyen2023vindr} 
        & Mammography datasets with scanner-specific biases; CTL enhances robustness across scanner variations. \\
        \midrule

        ACDC \cite{bernard2018deep}, Pancreas-CT \cite{clark2013cancer, roth2016data}, BraTS'19 \cite{bakas2017advancing} 
        & Multi-modality datasets for segmentation tasks, including cardiac, pancreas, and brain tumour imaging; CTL improves cross-modality generalisation. \\
        \midrule

        \href{http://brain-development.org/ixi-dataset/}{IXI}, fastMRI \cite{knoll2020fastmri} 
        & MRI datasets with multi-coil data across various contrasts and regions; CTL assesses domain shift handling for MRI reconstruction. \\
        \bottomrule
    \end{tabular}
\end{table*}

\section{Evaluating Causal Learning Models}

Evaluating causal learning models differs markedly from traditional metrics in image analysis, where the primary goal is typically to maximise predictive performance on tasks like classification, detection, or segmentation. In conventional image analysis, metrics such as accuracy, precision, recall, Intersection over Union (IoU), and mean Average Precision (mAP) assess how well a model identifies or localises patterns and objects in images based solely on associations in the data. However, these metrics focus on correlation rather than causation, meaning they are designed to optimise pattern recognition rather than uncover underlying mechanisms or causal relationships.

In contrast, causal models prioritise understanding cause-and-effect relationships within the data, shifting the focus away from prediction accuracy toward criteria that evaluate causal insights. As shown in Table \ref{tab:causal_evaluation}, causal models are assessed on their ability to infer structural and functional relationships through metrics like Intervention Testing, Counterfactual Reasoning, and Do-Calculus Validity. These metrics evaluate whether the model can accurately predict intervention outcomes or hypothetical changes in variables, and whether it adheres to causal rules that distinguish true causation from spurious correlation. For example, where image analysis would assess model performance based on how accurately a model segments an image, causal model evaluation might instead ask whether the model can predict the effect of modifying certain features on an outcome or if it can generalise across different contexts (as in Generalisation Under Distribution Shifts, see Table \ref{tab:causal_evaluation}). Causal models are therefore held to criteria that assess their robustness to distributional changes and their ability to reduce bias from confounding factors—capabilities critical for capturing true causal dynamics rather than merely recognising data patterns.

\begin{table*}[h]
\centering
\scriptsize
\caption{Evaluation Criteria for Causal Models}
\label{tab:causal_evaluation}
\begin{tabular}{@{}>{\raggedright\arraybackslash}p{3.5cm} 
                >{\raggedright\arraybackslash}p{9.5cm} 
                >{\centering\arraybackslash}p{2cm}@{}}
\toprule
\textbf{Evaluation Criterion} & \textbf{Description} & \textbf{Reference} \\
\midrule

\textbf{Intervention Testing} 
& A causal model should accurately predict the effects of interventions by testing how the model responds to changes in a variable. Comparison of its predictions to observed outcomes is often feasible through randomised controlled trials (RCTs) or synthetic experiments in simulation environments. 
& \cite{Peters2017, pearl2009causality} \\
\midrule

\textbf{Counterfactual Reasoning} 
& Counterfactual analysis assesses a model’s ability to predict hypothetical outcomes under different conditions, beyond observable data. Evaluation is challenging, but counterfactual accuracy can sometimes be tested in synthetic datasets with known counterfactual outcomes. 
& \cite{imbens2015causal} \\
\midrule

\textbf{Do-Calculus Validity} 
& For models based on SCMs and Judea Pearl’s causal framework, it is important to ensure compliance with the rules of do-calculus, which allows models to handle interventions by distinguishing causal from non-causal relationships. 
& \cite{pearl2009causality} \\
\midrule

\textbf{Synthetic Data with Known Causal Structures} 
& Evaluation on synthetic datasets with predefined causal structures helps in assessing causal discovery accuracy. Inferred causal structures can be compared with ground truth as a benchmark for validation. 
& \cite{Spirtes2000} \\
\midrule

\textbf{Benchmarking on Causal Datasets} 
& Certain datasets, such as the Twins and IHDP datasets in healthcare, are designed for causal inference tasks. These benchmarks enable evaluation of the model’s ability to capture treatment effects, often measured using metrics such as Average Treatment Effect (ATE) and Conditional Average Treatment Effect (CATE). 
& \cite{Hill2011} \\
\midrule

\textbf{Structural Metrics} 
& Causal discovery models that infer causal graphs can be evaluated using structural metrics, such as the Structural Hamming Distance (SHD), which assesses structural accuracy by counting the edits required to match the true causal graph. 
& \cite{Kalisch2012} \\
\midrule

\textbf{Generalisation Under Distribution Shifts} 
& Causal models are evaluated for robustness to distribution shifts to test if they generalise to out-of-distribution data, as causal relationships should remain consistent across different settings. 
& \cite{rojas2018invariant} \\
\midrule

\textbf{Bias Reduction and Confounder Control} 
& Many causal models aim to reduce biases from confounding variables. Success is measured by lower bias estimates than those of non-causal models, particularly for estimating treatment effects in observational data. 
& \cite{rubin1974estimating} \\
\bottomrule

\end{tabular}
\end{table*}

\section{Beyond Causal AI}

Causal AI represents a paradigm shift from association to causation in artificial intelligence, enabling more adaptive, interactive, and autonomous systems in healthcare. Future directions include adaptive AI that can self-calibrate and respond to patient variability in real time, causal discovery and counterfactual inference for understanding complex clinical relationships, and decision-centric models that tailor interventions to individual patients under uncertainty. Multimodal approaches that integrate medical images, clinical reports, and genomic data can improve diagnostic accuracy. In contrast, explainable, ethically aligned causal models enhance transparency, trust, and bias mitigation in medical decision-making. Additionally, neuro-symbolic causal AI combines domain knowledge with neural networks to produce scientifically grounded predictions, such as in drug discovery, and deeper clinician-AI collaboration enables interactive systems in which human expertise guides causal reasoning and treatment planning. 

\section{Conclusion and Discussion}

Deep learning has largely solved perception but not reliability. Most clinical failures of medical AI arise not from insufficient accuracy but from the breakdown of learned correlations under changing clinical conditions. Causal Transfer Learning provides the mathematical foundation for moving from correlation-driven models to mechanism-driven intelligence. By explicitly modelling causal structure, CTL enables systems to generalise across hospitals, populations, and technologies while remaining interpretable and fair. As medical AI moves from laboratory benchmarks to real-world deployment, CTL is poised to become a central pillar of trustworthy clinical machine learning.

CTL represents a transformative paradigm in medical image analysis, significantly enhancing the ability to analyse complex datasets and derive meaningful insights. By embedding causal principles into transfer learning methodologies, CTL offers a robust framework for addressing fundamental challenges, including domain adaptation, causal inference, and counterfactual reasoning. This capability is especially pertinent in medical image analysis, where the stakes are high, and the need for accurate diagnostics and effective treatment strategies is paramount.

Throughout this survey, we have explored various applications of CTL, including image classification, segmentation, predictive modelling, and multimodal data integration. Each application demonstrates the value of understanding the causal relationships inherent in medical datasets. For example, in image classification tasks, CTL has shown promise in distinguishing among various conditions using causal insights, thereby improving diagnostic accuracy. In segmentation tasks, CTL enhances the precision of delineating structures, such as tumours, thereby supporting more informed clinical decision-making. In addition, by incorporating multimodal data, including electronic health records and genomic information, CTL can develop comprehensive models that account for a wide range of factors influencing patient outcomes, thereby enabling a more holistic approach to healthcare.

Despite promising advances, several challenges and limitations remain to the broader adoption of CTL in clinical practice. Scalability is a significant concern, as the computational complexity of causal discovery and counterfactual inference can hinder real-time applications in dynamic healthcare settings. Furthermore, clinical validation of CTL models is crucial, as models that perform well in controlled research settings may not necessarily translate to success in real-world scenarios. Variations in patient demographics, imaging protocols, and clinical practices can significantly impact model performance, underscoring the need for comprehensive validation studies in diverse clinical contexts.

Ethical considerations also play a vital role in the development and implementation of CTL models. The potential for biased outcomes due to data imbalances or flawed causal assumptions could exacerbate existing disparities in healthcare. Researchers must proactively address these ethical implications, ensuring that CTL models are designed with fairness and transparency in mind. By fostering collaboration between data scientists, clinicians, and ethicists, the medical community can develop frameworks that promote equity in CTL applications.

Interpretability and explainability are paramount in medical AI applications, as clinicians must trust and understand the decisions made by these models. Although causal models provide a valuable structure for elucidating relationships among variables, ensuring these insights are communicated effectively to healthcare professionals remains a challenge. Further research is needed to develop interpretable CTL models that convey causal relationships in accessible, clinically relevant terms.

Although current CTL methods demonstrate promising robustness and generalisation capabilities in practical settings, several foundational challenges remain unresolved. Beyond the practical issues of scalability, clinical validation, ethics, and interpretability discussed above, there are also deeper theoretical limitations that warrant further investigation. In particular, future research must establish stronger theoretical guarantees regarding causal identifiability across heterogeneous clinical environments and determine whether learned latent representations genuinely capture causal mechanisms rather than stable statistical regularities. Furthermore, new evaluation paradigms are needed to assess intervention robustness, counterfactual consistency, and causal validity beyond conventional predictive metrics. Another important direction involves balancing domain invariance with the preservation of clinically meaningful population-specific variability, especially in diverse patient cohorts and multimodal imaging settings. Addressing these challenges will be critical for translating CTL into clinically trustworthy and deployable healthcare systems.

In conclusion, causal transfer learning holds significant promise for transforming medical image analysis by shifting the focus from correlation-based prediction to mechanism-aware modelling. By explicitly incorporating causal structure, CTL provides a principled pathway toward improved generalisation, interpretability, and fairness across heterogeneous clinical settings. However, realising this potential will require not only methodological advances but also careful attention to validation, deployment constraints, and clinical integration. With continued interdisciplinary effort, CTL has the potential to contribute to more reliable diagnostics, improved treatment strategies, and more equitable healthcare delivery.

\bibliographystyle{elsarticle-num} 
\bibliography{Mendeley}




\end{document}